\documentclass[journal]{IEEEtran}

\usepackage[numbers,sort&compress]{natbib}
\usepackage{amsmath,graphicx}
\usepackage{amsfonts}
\usepackage{multirow}
\usepackage{booktabs}
\usepackage{bm}
\usepackage{mathrsfs}
\usepackage{algorithm}
\usepackage{algorithmic}
\usepackage{setspace}
\usepackage{subfigure}
\usepackage[justification=centering]{caption}
\usepackage{float}
\usepackage{color}
\usepackage{makecell}
\usepackage{balance}

\newtheorem{theorem}{Theorem}
\newcommand{\reffig}[1]{Fig. \ref{#1}}

\begin{document}

\title{\huge {Multi-Kernel Correntropy for Robust Learning}}

\author{Badong~Chen,~\IEEEmembership{Senior Member,~IEEE,}
        Yuqing~Xie,
        Xin~Wang,~\IEEEmembership{Student Member,~IEEE,}
        Zejian~yuan,~\IEEEmembership{Member,~IEEE,}
        Pengju~Ren,~\IEEEmembership{Member,~IEEE,}
        and~Jing~Qin,~\IEEEmembership{Member,~IEEE}
\thanks{This work was supported by the National NSF of China (No. 61976175, No. 91648208), the National Natural Science Foundation-Shenzhen Joint Research Program (No.U1613219) and the National Key R$\&$D Program of China (2017YFB1002501).}
\thanks{Badong~Chen (Corresponding Author), Yuqing~Xie, Xin~Wang, Zejian~Yuan and Pengju~Ren(chenbd@mail.xjtu.edu.cn, felixxyq@stu.xjtu.edu.cn, wangxin0420@stu.xjtu.edu.cn, yuan.ze.jian@mail.xjtu.edu.cn and pengjuren@gmail.com) are with the Institute of Artificial Intelligence and Robotics,Xi'an Jiaotong University, Xi'an 710049, Shaanxi, China.}
\thanks{Jing~Qin(harry.qin@polyu.edu.hk) is with the Center of Smart Health, School of Nursing, Hong Kong Polytechnic University, Hongkong, China.}
}

\maketitle

\begin{abstract}
 As a novel similarity measure that is defined as the expectation of a kernel function between two random variables, correntropy has been successfully applied in robust machine learning and signal processing to combat large outliers. The kernel function in correntropy is usually a zero-mean Gaussian kernel. In a recent work, the concept of mixture correntropy (MC) was proposed to improve the learning performance, where the kernel function is a mixture Gaussian kernel, namely a linear combination of several zero-mean Gaussian kernels with different widths. In both correntropy and mixture correntropy, the center of the kernel function is, however, always located at zero. In the present work, to further improve the learning performance, we propose the concept of multi-kernel correntropy (MKC), in which each component of the mixture Gaussian kernel can be centered at a different location. The properties of the MKC are investigated and an efficient approach is proposed to determine the free parameters in MKC. Experimental results show that the learning algorithms under the maximum multi-kernel correntropy criterion (MMKCC) can outperform those under the original maximum correntropy criterion (MCC) and the maximum mixture correntropy criterion (MMCC).
\end{abstract}

\begin{IEEEkeywords}
Correntropy, mixture correntropy, multi-kernel correntropy, robust learning, outliers.
\end{IEEEkeywords}

\section{Introduction}
\IEEEPARstart{A} key problem in supervised machine learning is how to define an objective function to measure the similarity between model output and a target variable. The mean square error (MSE) is one of the most popular similarity measures, which is computationally simple and easy to use as a performance index in many signal processing and machine learning applications. The MSE is, however, vulnerable to non-Gaussian noises, such as impulsive noises or outliers, because the solution that minimizes the squared difference (the error in ${L_2}$ norm) can deviate far from the optimal solution in the presence of large outliers. To address this problem, many non-MSE similarity measures were proposed in the literature, such as the mean absolute error (MAE)\cite{lin1990adaptive,coyle1988stack}, mean p-power error (MPE)\cite{2pei1994least}, M-estimation cost \cite{liu2006error} and logarithmic cost \cite{sayin2014novel}. In particular in recent years, the correntropy as a local similarity measure in kernel space has found many successful applications in robust regression \cite{chen2012recursive,13feng2015learning}, classification \cite{xu2018robust,syed2012correntropy,singh2014c,shi2018training,ren2020correntropy}, PCA \cite{he2011robust}, feature extraction \cite{zhou2017maximum,yu2020correntropy}, adaptive filtering \cite{zhao2011kernel,chen2016generalized,ma2015maximum,wu2015robust,wu2015kernel,cinar2012hidden,liu2017maximum} and so on. Correntropy defines a non-homogeneous metric (Correntropy Induced Metric, CIM) that behaves like different norms (from ${L_2}$ to ${L_0}$) depending on the actual distance between samples, which can be used as an outlier-robust error measure in robust signal processing or a sparsity penalty term in sparse signal processing \cite{liu2007correntropy}.

The original correntropy is defined as the expectation of a kernel function between two random variables, where the kernel function is usually a zero-mean Gaussian kernel \cite{liu2007correntropy}. The learning methods under maximum correntropy criterion (MCC) may, however, perform poorly when the kernel function in correntropy is limited to a single Gaussian kernel. It is likely that the combination of several kernel functions can perform much better. The mixture correntropy (MC) was thus proposed in a recent work to improve the learning performance, in which the kernel function is implemented by a linear combination of several zero-mean Gaussian kernels with different widths \cite{chen2018mixture}. Similar ideas can be found in multiple kernel learning (MKL) methods \cite{gonen2011multiple}, such as the Multiple Kernel Support Vector Machine (MKSVM) \cite{lanckriet2004learning}, Multiple Kernel Modification of Ho-Kashyap algorithm with Squared approximation of the misclassification errors (MultiK-MHKS) \cite{wang2007multik} and Multikernel Adaptive Filtering (MKAF)\cite{yukawa2012multikernel}, where a combination of several kernels is used instead of a single kernel. However, there is still a shortcoming in the mixture correntropy that only allows the combination of zero-mean Gaussian kernels, which may perform poorly under some complex non-Gaussian noises such as those from multimodal distributions. To further improve the learning performance, in the present work, we propose a novel concept of multi-kernel correntropy (MKC), where each component of the mixture Gaussian kernel can be centered at a different location (not limited to zero-mean). Some important properties of the MKC are also studied. The MKC involves more free parameters than the MC, so a challenging issue is how to determine the free parameters in a practical application. To address this issue, we propose an efficient approach in this paper to optimize the free parameters in MKC by minimizing a distance between the mixture Gaussian function and the error's probability density function (PDF). Experimental results have confirmed the satisfactory performance of the learning methods under maximum multi-kernel correntropy criterion (MMKCC). Due to its excellent flexibility and robustness, the proposed MKC has great potential to be applied in many fields involving complex noise disturbances, such as biomedical engineering, remote sensing, autonomous systems and many others.

The rest of the paper is organized as follows. In section II, we define the MKC and present several properties. In section III, we propose an effective method to optimize the free parameters in MKC. Experimental results are then presented in section IV and finally, conclusion is given in section V.

\section{MULTI-KERNEL CORRENTROPY}
\subsection{Definitions}
Given two random variables $X \in \mathbf{R}$ and $Y \in \mathbf{R}$ with joint PDF ${p_{XY}}(x,y)$, correntropy is defined by \cite{liu2007correntropy}
\begin{equation}\label{eq1}
\begin{aligned}
  V(X,Y)=\mathbf{E}[\kappa(X,Y)]=\iint{\kappa(x,y)p_{XY}(x,y)dxdy}
\end{aligned}
\end{equation}
where $\kappa (.,.)$ is usually a radial kernel, and $\textbf{E}[.]$ denotes the expectation operator. If the kernel function $\kappa (.,.)$ satisfies Mercer's condition, correntropy can be expressed as a correlation measure in a functional Hilbert space $\mathcal{F}$:
\begin{equation}
\label{eq2}
\begin{aligned}
V(X,Y) = \rm{\textbf{E}}\left[ {{{\left\langle {\varphi (\emph{X}),\varphi (\emph{Y})} \right\rangle }_\mathcal{F}}} \right]
\end{aligned}
\end{equation}
where $\varphi (.)$ is a nonlinear mapping induced by the kernel to transform the variables from the original space to the functional space $\mathcal{F}$, and ${\left\langle {.,.} \right\rangle _\mathcal{F}}$ stands for the inner product in $\mathcal{F}$. Without explicit mention, the kernel function in correntropy is the well-known Gaussian kernel:
\begin{equation}
\label{eq3}
\begin{aligned}
\kappa (X,Y) = {\kappa _\sigma }(e) = \frac{1}{{\sqrt {2\pi } \sigma }}\exp \left( { - \frac{{{e^2}}}{{2{\sigma ^2}}}} \right)
\end{aligned}
\end{equation}
where $e = X - Y$ is the error between $X$ and $Y$, and $\sigma$ is the kernel bandwidth ($\sigma>0$). It is easy to understand that correntropy measures how similar two random variables are in a local region of the error space controlled by the kernel bandwidth. Correntropy can easily be estimated from finite samples:
\begin{equation}
\label{eq4}
\begin{aligned}
{\hat V_\sigma }(X,Y) = \frac{1}{N}\sum\limits_{i = 1}^N {{\kappa _\sigma }({x_i} - {y_i})}
\end{aligned}
\end{equation}
where $\left\{ {{x_i},{y_i}} \right\}_{i = 1}^N$ are $N$ samples of the random variables $X$ and $Y$. In particular, the function $CIM(\tilde X,\tilde Y) = \sqrt {{\kappa _\sigma }(0) - {{\hat V}_\sigma }(X,Y)}$ defines a metric, namely the correntropy induced metric (CIM) in the sample space, where $\tilde X = {\left[ {{x_1}, \cdots ,{x_N}} \right]^T}$, $\tilde Y = {\left[ {{y_1}, \cdots ,{y_N}} \right]^T}$. The CIM behaves like an $L_2$ norm distance if samples are close and like an $L_1$ norm distance as samples get further apart and eventually will approach the $L_0$ norm as samples far apart. This property elucidates the robustness of correntropy for outlier rejection. Under the maximum correntropy criterion (MCC), the detrimental effect of outliers can effectively be eliminated by maximizing the correntropy between the model output and target response \cite{chen2017maximum}.

The kernel function in correntropy is usually limited to a zero-mean Gaussian kernel and this may seriously restricts its performance when used as a cost function in machine learning. To improve the learning performance, the mixture correntropy (MC) was proposed in a recent paper \cite{chen2018mixture} by using a linear combination of several zero-mean Gaussian kernels (with different bandwidths) as the kernel function. The mixture correntropy with $m$ sub-kernels is
\begin{equation}\small
\label{eq5}
\begin{aligned}
{V_{\lambda ,\sigma }}(X,Y) &= \sum\limits_{i = 1}^m {{\lambda _i}{V_{{\sigma _i}}}(X,Y)} \\
&= \rm{\textbf{E}}\Big[ {\sum\nolimits_{i = 1}^m {{\lambda _i}{\kappa _{{\sigma _i}}}(\emph{X - Y})} } \Big]\\
&= \iint{{\Big( {\sum\nolimits_{i = 1}^m {{\lambda _i}{\kappa _{{\sigma _i}}}(x - y)} } \Big)}p_{XY}}(x,y)dxdy
\end{aligned}
\end{equation}
where $\bm{\lambda}  = {\left[ {{\lambda _1},{\lambda _2}, \cdots ,{\lambda _m}} \right]^T}$ is the mixture coefficient vector, and $\bm{\sigma}  = {\left[ {{\sigma _1},{\sigma _2}, \cdots ,{\sigma _m}} \right]^T}$ is the bandwidth vector. Usually, the mixture coefficient vector satisfies $\sum\limits_{i = 1}^m {{\lambda _i}}  = 1$ with ${\lambda _i} \ge 0$($i = 1, \cdots ,m$). In \cite{chen2018mixture}, for simplicity, only the case of $m=2$ is considered. There is still a limitation in the mixture correntropy, that is, all the sub-kernels are centered at zero. To solve this limitation and further enhance the learning performance, in the present paper, we propose a more general definition of correntropy, namely, the multi-kernel correntropy (MKC), in which the sub-kernels can be centered at different locations (not limited to zero-mean). Specifically, the MKC between random variables $X$ and $Y$ is defined by
\begin{equation}
\label{eq6}
\begin{aligned}
&{V_{\bm{\lambda ,c,\sigma} }}(X,Y)= \textbf{E}\left[ {\sum\limits_{i = 1}^m {{\lambda _i}{\kappa _{{\sigma _i}}}(X - Y - {c_i})} } \right]\\
&=\iint{{\left( {\sum\limits_{i = 1}^m {{\lambda _i}{\kappa _{{\sigma _i}}}(x - y - {c_i})} } \right)}p_{XY}}(x,y)dxdy
\end{aligned}
\end{equation}
where $\bm{c} = {\left[ {{c_1},{c_2}, \cdots ,{c_m}} \right]^T} \in {\mathbf{R}^m}$ is the center vector.

\emph{Remark}: The kernel function in the above MKC is a multi-Gaussian function that usually does not satisfy Mercer's condition. This is not a problem, however, because for a similarity measure the Mercer's condition is not necessary.

\begin{figure}[]
	\setlength{\abovecaptionskip}{0pt}
	\setlength{\belowcaptionskip}{0pt}
	\centering
    \includegraphics[height=1.8in]{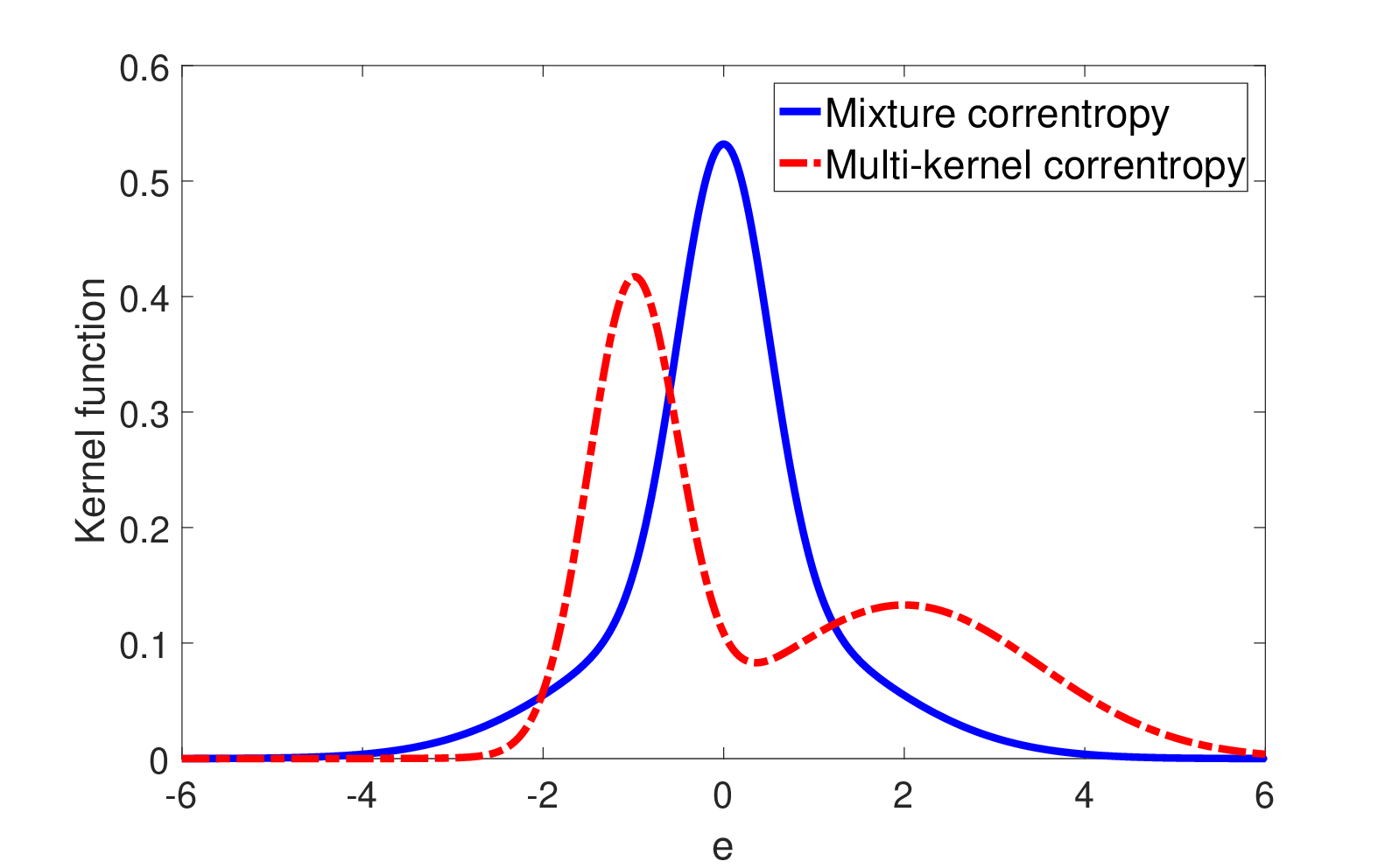}
	\caption{Kernel functions of the mixture correntropy and multi-kernel correntropy}
	\label{fig1}
\end{figure}

The MKC ${V_{\bm{\lambda ,c,\sigma} }}(X,Y)$ will reduce to the MC ${V_{\bm{\lambda ,\sigma} }}(X,Y)$ when $\bm{c} = {\left[ {0, \cdots ,0} \right]^T}$. \reffig{fig1} shows the kernel functions of the mixture correntropy($m=2, \lambda_1=0.5, \lambda_2=0.5, \sigma_1=0.5, \sigma_2=1.5$) and multi-kernel correntropy($m=2, \lambda_1=0.5, \lambda_2=0.5, \sigma_1=0.5, \sigma_2=1.5, c_1=-1.0, c_2=2.0$). Compared with the MC, the MKC is much more general and flexible and can adapt to more complicated error distribution, such as skewed, multi-peak, discrete-valued distribution, and hence it may achieve much better performance with proper setting of the centers when used as a cost function in machine learning. However, the MKC contains  $3m$ free parameters, which have to be determined in practical applications. We will develop an efficient method in section IV to determine these free parameters.
\subsection{Properties}
In the following, we present several properties of the MKC. The first and second properties are very straightforward and will not be proved here.

Property 1:  The MKC ${V_{\bm{\lambda ,c,\sigma} }}(X,Y)$ is positive and bounded: $0 < {V_{\bm{\lambda ,c,\sigma} }}(X,Y) \le \sum\limits_{i = 1}^m {\frac{{{\lambda _i}}}{{\sqrt {2\pi } {\sigma _i}}}} $

Property 2: The MKC ${V_{\bm{\lambda ,c,\sigma} }}(X,Y)$  involves all the even moments of the error $e = X - Y$ about the centers $\left\{ {{c_i}} \right\}$, that is,
\begin{small}
\begin{equation}
\label{eq7}
\begin{aligned}
{V_{\bm{\lambda ,c,\sigma} }}(X,Y) = \sum\limits_{i = 1}^m {\left( {\frac{{{\lambda _i}}}{{\sqrt {2\pi } {\sigma _i}}}\sum\limits_{n = 0}^\infty  {\frac{{{{\left( { - 1} \right)}^n}}}{{{2^n}n!}}\textbf{E}\left[ {\frac{{{{\left( {e - {c_i}} \right)}^{2n}}}}{{\sigma _i^{2n}}}} \right]} } \right)}
\end{aligned}
\end{equation}
\end{small}

\emph{Remark}: As $\left\{ {{\sigma _i}} \right\}$ increases, the high-order moments will decay fast, and the second-order moments will tend to dominate the value.

Property 3: As $min\left\{ {{\sigma _i}} \right\}$ is large enough, it holds that
\begin{equation}
\label{eq8}
\begin{aligned}
{V_{\bm{\lambda ,c,\sigma} }}(X,Y) \approx \sum\limits_{i = 1}^m {\frac{{{\lambda _i}}}{{\sqrt {2\pi } {\sigma _i}}}} \left( {1 - \frac{1}{{2\sigma _i^2}}\textbf{E}\left[ {{{\left( {e - {c_i}} \right)}^2}} \right]} \right)
\end{aligned}
\end{equation}

Proof:  Since $\exp (x) \approx 1 + x$ for $x$ small enough, as $min\left\{ {{\sigma _i}} \right\}$ is large enough, we have
\begin{equation}
\label{eq9}
\begin{aligned}
{V_{\bm{\lambda ,c,\sigma} }}(X,Y) &= \textbf{E}\left[ {\sum\limits_{i = 1}^m {{\lambda _i}{\kappa _{{\sigma _i}}}(X - Y - {c_i})} } \right]\\
{\rm{                 }} &\approx \textbf{E}\left[ {\sum\limits_{i = 1}^m {\frac{{{\lambda _i}}}{{\sqrt {2\pi } {\sigma _i}}}\left( {1 - \frac{{{{\left( {e - {c_i}} \right)}^2}}}{{2\sigma _i^2}}} \right)} } \right]\\
{\rm{                 }} &= \sum\limits_{i = 1}^m {\frac{{{\lambda _i}}}{{\sqrt {2\pi } {\sigma _i}}}} \left( {1 - \frac{1}{{2\sigma _i^2}}\textbf{E}\left[ {{{\left( {e - {c_i}} \right)}^2}} \right]} \right)
\end{aligned}
\end{equation}
which completes the proof.

\emph{Remark}: According to Property 3, when $min\left\{ {{\sigma _i}} \right\}$ is very large, maximizing the MKC will be equivalent to minimizing a weighted sum of the error's second-order moments about the centers $\left\{ {{c_i}} \right\}$.

Property 4: Let ${p_e}(.)$ be the PDF of the error variable $e=X-Y$. It holds that
\begin{equation}
\label{eq10}
\begin{aligned}
\mathop {\lim }\limits_{max\left\{ {{\sigma _i}} \right\} \to 0 + } {V_{\bm{\lambda ,c,\sigma} }}(X,Y) = \sum\limits_{i = 1}^m {{\lambda _i}{p_e}({c_i})}
\end{aligned}
\end{equation}

Proof:  When $max\left\{ {{\sigma _i}} \right\}$ shrinks to zero, the Gaussian kernel function ${\kappa _{{\sigma _i}}}(.)$ will approach the Dirac delta function $\delta (.)$. Thus we have
\begin{small}
\begin{equation}
\label{eq11}
\begin{aligned}
\mathop {\lim }\limits_{max\left\{ {{\sigma _i}} \right\} \to 0 + }&{V_{\bm{\lambda ,c,\sigma} }}(X,Y) = \mathop {\lim }\limits_{max\left\{ {{\sigma _i}} \right\} \to 0 + } \textbf{E}\left[ {\sum\limits_{i = 1}^m {{\lambda _i}{\kappa _{{\sigma _i}}}(e - {c_i})} } \right]\\
&= \mathop {\lim }\limits_{max\left\{ {{\sigma _i}} \right\} \to 0 + } \sum\limits_{i = 1}^m {{\lambda _i}\int {{\kappa _{{\sigma _i}}}(\varepsilon  - {c_i}){p_e}(\varepsilon )d\varepsilon } } \\
&= \sum\limits_{i = 1}^m {{\lambda _i}\int {\delta (\varepsilon  - {c_i}){p_e}(\varepsilon )d\varepsilon } } \\
&= \sum\limits_{i = 1}^m {{\lambda _i}{p_e}({c_i})}
\end{aligned}
\end{equation}
\end{small}
which completes the proof.

\emph{Remark}: According to Property 4, when $max\left\{ {{\sigma _i}} \right\}$ is very small, the MKC will approach a weighted sum of the values of ${p_e}(\varepsilon )$ evaluated at $\varepsilon  = {c_i}$ ($i = 1, \cdots ,m$).

\section{MAXIMUM MULTI-KERNEL CORRENTROPY CRITERION}
The proposed MKC can be used to build new cost functions in many machine learning applications. Consider a supervised learning setting where the goal is to optimize a model $M$ that receives a random variable $X$ and outputs $Y=M(X)$ that should approximate a target variable (or teaching variable) $T$. Here $M(.)$ denotes an unknown mapping from the input to output that needs to be learned. A central problem in this learning task is the definition of a loss function (or a similarity measure) to compare $Y$ with $T$. The well-known minimum mean square error (MMSE) criterion has been the workhorse of supervised learning, which aims to minimize the MSE cost $\textbf{E}\left[ {{e^2}} \right]$ with $e=T-Y$ being the error variable. The combination of the linear feedforward model and MSE yields a set of equations that can be solved analytically. However, MSE is only optimal when the error variable is Gaussian distributed, which is seldom the case in real world applications. The error distributions tend to be skewed and with long tails, which create problems for MSE. Therefore, many "optimal solutions" are indeed not practical, simply because of the criterion that is used in the optimization. Many non-MSE optimization criterion were proposed in the literature to address the limitations of the MSE. The maximum correntropy criterion (MCC) is one of the hotspots of current research, which performs very well particularly when the error distribution is heavy-tailed \cite{wang2017maximum}. Under the MCC, the model is optimized (or trained) to maximize the correntropy between the target $T$ and output $Y$:
\begin{equation}
\label{eq12}
\begin{aligned}
{M^*} &= \mathop {\arg \max }\limits_{M \in \textbf{M}} {V_\sigma }\left( {T,Y} \right) \\
&= \mathop {\arg \max }\limits_{M \in \textbf{M}} \textbf{E}\left[ {{\kappa _\sigma }\left( e \right)} \right]
\end{aligned}
\end{equation}
where $M^*$ denotes the optimal model and $\textbf{M}$ stands for the hypothesis space. To improve the learning performance, the maximum mixture correntropy criterion (MMCC) was proposed in \cite{chen2018mixture}. To further improve the flexibility and robustness, in the present paper, we propose the maximum multi-kernel correntropy criterion (MMKCC), where the optimal model is obtained by maximizing the MKC, that is
\begin{equation}
\label{eq13}
\begin{aligned}
{M^*} &= \mathop {\arg \max }\limits_{M \in \textbf{M}} {V_{\bm{\lambda ,c,\sigma} }}(T,Y) \\
&= \mathop {\arg \max }\limits_{M \in \textbf{M}} \textbf{E}\left[ {\sum\limits_{i = 1}^m {{\lambda _i}{\kappa _{{\sigma _i}}}(e - {c_i})} } \right]
\end{aligned}
\end{equation}
In a practical situation, given finite input-target samples $\left\{ {{x_j},{t_j}} \right\}_{j = 1}^N$, the model can be trained through maximizing a sample estimator of the MKC:
\begin{equation}
\label{eq14}
\begin{aligned}
{M^*} &= \mathop {\arg \max }\limits_{M \in \textbf{M}} {\hat V_{\bm{\lambda ,c,\sigma} }}(T,Y) \\
&= \mathop {\arg \max }\limits_{M \in \textbf{M}} \frac{1}{N}\sum\limits_{j = 1}^N {\sum\limits_{i = 1}^m {{\lambda _i}{\kappa _{{\sigma _i}}}({e_j} - {c_i})} }
\end{aligned}
\end{equation}
where ${e_j} = {t_j} - {y_j} = {t_j} - M({x_j})$ is the  $j$-th error sample.

In the following, we present a simple example to show how to solve the optimal solution under MMKCC. Consider a linear-in-parameter (LIP) model in which the $j$-th output sample is
\begin{equation}
\label{eq15}
\begin{aligned}
{y_j} &= {\bm{h}_j}\bm{\beta}  \\
&= \left[ {{\varphi _1}({\bm{x}_j}),{\varphi _2}({\bm{x}_j}), \cdots ,{\varphi _L}({\bm{x}_j})} \right]{\left[ {{\beta _1},{\beta _2}, \cdots ,{\beta _L}} \right]^T}
\end{aligned}
\end{equation}
where ${\bm{h}_j} = \left[ {{\varphi _1}({\bm{x}_j}),{\varphi _2}({\bm{x}_j}), \cdots ,{\varphi _L}({\bm{x}_j})} \right] \in {\rm{\textbf{R}}^L}$ is the $j$-th nonlinearly mapped input vector (a row vector), with ${\varphi _l}(.)$ being the $l$-th nonlinear mapping function ($l = 1,2, \cdots ,L$), and $\bm{\beta}  = {\left[ {{\beta _1},{\beta _2}, \cdots ,{\beta _L}} \right]^T} \in {\rm{\textbf{R}}^L}$ is the output weight vector to be learned. Based on the MMKCC, the optimal weight vector ${\bm{\beta} ^*}$ can be solved by maximizing the following objective function:
\begin{equation}
\label{eq16}
\begin{aligned}
{\bm{\beta} ^*} &= \mathop {\arg \max }\limits_{\bm{\beta}  \in {\rm{\textbf{R}}^L}} J(\bm{\beta} ) \\
&= \frac{1}{N}\sum\limits_{j = 1}^N {\sum\limits_{i = 1}^m {{\lambda _i}{\kappa _{{\sigma _i}}}({e_j} - {c_i})} }  - \gamma {\left\| \bm{\beta}  \right\|^2}
\end{aligned}
\end{equation}
where ${e_j} = {t_j} - {\bm{h}_j}\bm{\beta} $ , and $\gamma  \ge 0$ is a regularization parameter. Setting ${{\partial J(\bm{\beta} )} \mathord{\left/
 {\vphantom {{\partial J(\bm{\beta} )} {\partial \bm{\beta} }}} \right.
 \kern-\nulldelimiterspace} {\partial \beta }} = \bm{0}$, we have
\begin{footnotesize}
\begin{equation}
\label{eq17}
\begin{aligned}
&\frac{1}{N}\sum\limits_{j = 1}^N {\sum\limits_{i = 1}^m {\frac{{{\lambda _i}}}{{\sigma _i^2}}{\kappa _{{\sigma _i}}}({e_j} - {c_i})({e_j} - {c_i})\bm{h}_j^T} }  - 2\gamma \bm{\beta}  = 0\\
 &\Rightarrow \sum\limits_{j = 1}^N {\sum\limits_{i = 1}^m {\frac{{{\lambda _i}}}{{\sigma _i^2}}{\kappa _{{\sigma _i}}}({e_j} - {c_i})({t_j} - {\bm{h}_j}\bm{\beta}  - {c_i})\bm{h}_j^T} }  - \gamma '\bm{\beta}  = 0\\
 &\Rightarrow \sum\limits_{j = 1}^N {\psi ({e_j})\bm{h}_j^T{\bm{h}_j}\beta }  + \gamma '\bm{\beta}  = \sum\limits_{j = 1}^N {\psi ({e_j}){t_j}\bm{h}_j^T}  - \sum\limits_{j = 1}^N {\zeta ({e_j})\bm{h}_j^T}
\end{aligned}
\end{equation}
\end{footnotesize}
where $\psi ({e_j}) = \sum\limits_{i = 1}^m {\frac{{{\lambda _i}}}{{\sigma _i^2}}{\kappa _{{\sigma _i}}}({e_j} - {c_i})}$, $\zeta ({e_j}) = \sum\limits_{i = 1}^m {\frac{{{\lambda _i}{c_i}}}{{\sigma _i^2}}{\kappa _{{\sigma _i}}}({e_j} - {c_i})}$, and $\gamma ' = 2N\gamma$. From (17), one can easily derive
\begin{equation}
\footnotesize
\label{eq18}
\begin{aligned}
\bm{\beta}  &= {\left( {\sum\limits_{j = 1}^N {\psi ({e_j})\bm{h}_j^T{\bm{h}_j}}  + \gamma 'I} \right)^{ - 1}}\left( {\sum\limits_{j = 1}^N {\psi ({e_j}){t_j}\bm{h}_j^T}  - \sum\limits_{j = 1}^N {\zeta ({e_j})\bm{h}_j^T} } \right)\\
&= {\left( {{\textbf{H}^T}\bm{\Lambda} \textbf{H} + \gamma 'I} \right)^{ - 1}}\left( {{\textbf{H}^T}\bm{\Lambda} \textbf{T} - {\textbf{H}^T}\bm{\theta} } \right)
\end{aligned}
\end{equation}
where $\textbf{H} = \left[ {{\bm{h}_{jl}}} \right]$ is an $N \times L$ dimensional matrix with ${\bm{h}_{jl}} = {\varphi _l}({\bm{x}_j})$, $\bm{\Lambda}$ is an $N \times N$ diagonal matrix with diagonal elements ${\bm{\Lambda}_{jj}} = \psi ({e_j})$, $\textbf{T} = {\left[ {{t_1}, \cdots ,{t_N}} \right]^T}$ , and $\bm{\theta}  = {\left[ {\zeta ({e_1}), \cdots ,\zeta ({e_N})} \right]^T}$.

The equation (18) is not a closed-form solution and it is actually a fixed-point equation because the diagonal matrix $\bm{\Lambda}$ and vector $\bm{\theta}$ on the right-hand side depend on the weight vector $\bm{\beta}$ through ${e_j} = {t_j} - {\bm{h}_j}\bm{\beta} $. Thus, the optimal solution of $\bm{\beta}$ can be obtained via a fixed-point iterative algorithm under MMKCC (FP-MMKCC), as described in \textbf{Algorithm 1}.

The computational complexities of some steps are given in Table I. Then, the computational complexity of the FP-MMKCC algorithm is $\big[2L^2N+8LN+21mN-2N-L^2+O(L^3)\big]T_\text{FP}$, where $T_\text{FP}$ is the fixed-point iteration number. Since the fixed-point iteration number $T_\text{FP}$ is relatively small in general, the computational complexity of the FP-MMKCC algorithm is moderate. Moreover, a sufficient condition to guarantee the convergence of the FP-MMKCC algorithm can be obtained (See APPENDIX A).

\begin{algorithm}
	\renewcommand{\algorithmicrequire}{\textbf{Input:}}
	\renewcommand{\algorithmicensure}{\textbf{Output:}}
	\caption{FP-MMKCC algorithm}
	\label{alg:1}
	\begin{algorithmic}[1]
		\REQUIRE training samples $\{ {\bm{x}_i},{t_i}\} _{i = 1}^N$, number of nonlinear mappers $L$,  mixture coefficient vector $\bm{\lambda}$, bandwidth vector $\bm{\sigma}$, center vector $\bm{c}$, regularization parameter $\gamma'$, maximum iteration number $K$, termination tolerance $\xi$ and the initial weight vector ${\bm{\beta} _0}{\rm{ = }}\textbf{0}$.
		\ENSURE weight vector $\bm{\beta}$\\
		\FORALL{$k = 1,2,...,K$}
		\STATE Compute the errors based on ${\bm{\beta} _{k - 1}}$: ${e_i} = {t_i} - {\bm{h}_i}{\bm{\beta} _{k - 1}}$, $i = 1,2, \cdots ,N$
		\STATE Compute the diagonal matrix ${\bf{\Lambda }}$: ${\bm{\Lambda} _{jj}} = \sum\limits_{i = 1}^m {\frac{{{\lambda _i}}}{{\sigma _i^2}}{\kappa _{{\sigma _i}}}({e_j} - {c_i})}$,$j = 1,2, \cdots ,N$
		\STATE Compute the vector $\bm{\theta}$: $\bm{\theta}  = {\left[ {\zeta ({e_1}), \cdots ,\zeta ({e_N})} \right]^T}$
		\STATE Update the weight vector $\bm{\beta}$: ${\bm{\beta} _k} = {\left( {{\bf{H}^T}\bm{\Lambda} \bf{H} + \gamma 'I} \right)^{ - 1}}\left( {{\bf{H}^T}\bm{\Lambda} \bf{T} - {\bf{H}^T}\bm{\theta} } \right)$
        \STATE \textbf{Until} $\left| {{J}({\bm{\beta} _k}) - {J}({\bm{\beta} _{k - 1}})} \right| < \xi $
		\ENDFOR
	\end{algorithmic}
\end{algorithm}

\begin{table}[!h]
\label{complexity}
\centering
\small
\caption{Computational complexity for each iteration of the FP-MMKCC algorithm }\label{complexities}
\begin{tabular}{ccc}
  \hline
  Step &  \makecell[c]{Addition/subtraction and\\multiplication} & \makecell[c]{Division, matrix inversion,\\ and exponentiation} \\
  \hline
  $2$ & $2LN$           & $0$\\
  $3$ & $5mN-N$         & $5mN$\\
  $4$ & $6mN-N$         & $5mN$\\
  $5$ & $2L^2N+6LN-L^2$ & $O(L^3)$\\
  \hline
\end{tabular}
\end{table}

\section{DETERMINATION OF FREE PARAMETERS IN MMKCC}
One of the most challenging problems in MMKCC is how to determine the $3m$ free parameters, namely the vectors $\bm{\lambda}  = {\left[ {{\lambda _1},{\lambda _2}, \cdots ,{\lambda _m}} \right]^T}$, $\bm{c} = {\left[ {{c_1},{c_2}, \cdots ,{c_m}} \right]^T}$ and $\bm{\sigma}  = {\left[ {{\sigma _1},{\sigma _2}, \cdots ,{\sigma _m}} \right]^T}$. If this problem is not solved, the MMKCC will not be practical. To address this problem, we consider again the supervised learning setting in the previous section. First, we divide the MMKCC into three terms:

\begin{equation}
\label{eq19}
\begin{aligned}
&{V_{\bm{\lambda ,c,\sigma} }}(T,Y) = \textbf{E}\left[ {\sum\limits_{i = 1}^m {{\lambda _i}{\kappa _{{\sigma _i}}}(e - {c_i})} } \right]\\
&=\frac{1}{2}\int {{{\left( {\sum\limits_{i = 1}^m {{\lambda _i}{\kappa _{{\sigma _i}}}(\varepsilon  - {c_i})} } \right)}^2}d\varepsilon } {\rm{ + }}\frac{1}{2}\int {{{\left( {{p_e}(\varepsilon )} \right)}^2}d\varepsilon } \\
&- \frac{1}{2}\int {{{\left( {\sum\limits_{i = 1}^m {{\lambda _i}{\kappa _{{\sigma _i}}}(\varepsilon  - {c_i})}  - {p_e}(\varepsilon )} \right)}^2}d\varepsilon }
\end{aligned}
\end{equation}

The first term is independent of the model $M$, so we have
\begin{equation}
\label{eq20}
\begin{aligned}
{M^*} &= \mathop {\arg \max }\limits_{M \in \textbf{M}} {V_{\bm{\lambda ,c,\sigma} }}(T,Y) \\
&= \mathop {\arg \max }\limits_{M \in \textbf{M}} {U_{\bm{\lambda ,c,\sigma} }}(T,Y)
\end{aligned}
\end{equation}
where ${U_{\bm{\lambda ,c,\sigma} }}(T,Y) = \frac{1}{2}\int {{{\left( {{p_e}(\varepsilon )} \right)}^2}d\varepsilon }  - \frac{1}{2}\int {{{\left( {\sum\limits_{i = 1}^m {{\lambda _i}{\kappa _{{\sigma _i}}}(\varepsilon  - {c_i})}  - {p_e}(\varepsilon )} \right)}^2}d\varepsilon }$.

\begin{figure*}
\begin{align*}\label{eq21}
&\left( {{M^*},\bm{{\lambda ^*},{c^*},{\sigma ^*}}} \right) = \mathop {\arg \max }\limits_{M \in \textbf{M},\bm{\lambda}  \in {\bm{\Omega _\lambda} },\bm{c} \in {\bm{\Omega _c}},\bm{\sigma}  \in {\bm{\Omega _\sigma} }} {U_{\bm{\lambda ,c,\sigma} }}(T,Y)\\
&= \mathop {\arg \max }\limits_{M \in \textbf{M},\bm{\lambda}  \in {\bm{\Omega _\lambda} },\bm{c} \in {\bm{\Omega _c}},\bm{\sigma}  \in {\bm{\Omega _\sigma} }} \frac{1}{2}\int {{{\left( {{p_e}(\varepsilon )} \right)}^2}d\varepsilon }  - \frac{1}{2}\int {{{\left( {\sum\limits_{i = 1}^m {{\lambda _i}{\kappa _{{\sigma _i}}}(\varepsilon  - {c_i})}  - {p_e}(\varepsilon )} \right)}^2}d\varepsilon } \\
&= \mathop {\arg \max }\limits_{M \in \textbf{M},\bm{\lambda}  \in {\bm{\Omega _\lambda} },\bm{c} \in {\bm{\Omega _c}},\bm{\sigma}  \in {\bm{\Omega _\sigma} }}  - \frac{1}{2}\int {{{\left( {\sum\limits_{i = 1}^m {{\lambda _i}{\kappa _{{\sigma _i}}}(\varepsilon  - {c_i})} } \right)}^2}d\varepsilon }  + \textbf{E}\left[ {\sum\limits_{i = 1}^m {{\lambda _i}{\kappa _{{\sigma _i}}}(e - {c_i})} } \right]
\tag{21}
\end{align*}
\begin{align*}\label{eq23}
\left( {{M^*},{\bm{\lambda ^*},{c^*},{\sigma ^*}}} \right) &= \mathop {\arg \max }\limits_{M \in \textbf{M},\bm{\lambda  \in {\Omega _\lambda },c \in {\Omega _c},\sigma  \in {\Omega _\sigma }}}  - \frac{1}{2}{\bm{\lambda} ^T}\left( {\int {\bm{\tilde g}(\varepsilon )\bm{\tilde g}{{(\varepsilon )}^T}} d\varepsilon } \right)\bm{\lambda}{\rm{ + }}{\bm{\lambda} ^T}\bm{\tilde h}\\
&= \mathop {\arg \max }\limits_{M \in \textbf{M},\bm{\lambda  \in {\Omega _\lambda },c \in {\Omega _c},\sigma  \in {\Omega _\sigma }}}  - \frac{1}{2}{\bm{\lambda} ^T}\mathbf{\tilde K}\bm{\lambda} {\rm{ + }}{\bm{\lambda} ^T} \bm{\tilde h}
\tag{23}
\end{align*}
\begin{align*}\label{eq24}
\bf{\tilde K} = \left[ {\begin{array}{*{20}{c}}
{\frac{1}{{\sqrt {2\pi } \sqrt {{\sigma _1}^2 + {\sigma _1}^2} }}\exp (-\frac{{{{({c_1} - {c_1})}^2}}}{{2({\sigma _1}^2 + {\sigma _1}^2)}})}& \ldots &{\frac{1}{{\sqrt {2\pi } \sqrt {{\sigma _1}^2 + {\sigma _m}^2} }}\exp (-\frac{{{{({c_1} - {c_m})}^2}}}{{2({\sigma _1}^2 + {\sigma _m}^2)}})}\\
 \vdots & \ddots & \vdots \\
{\frac{1}{{\sqrt {2\pi } \sqrt {{\sigma _m}^2 + {\sigma _1}^2} }}\exp (-\frac{{{{({c_m} - {c_1})}^2}}}{{2({\sigma _m}^2 + {\sigma _1}^2)}})}& \ldots &{\frac{1}{{\sqrt {2\pi } \sqrt {{\sigma _{m}}^2 + {\sigma _m}^2} }}\exp (-\frac{{{{({c_m} - {c_m})}^2}}}{{2({\sigma _m}^2 + {\sigma _m}^2)}})}
\end{array}} \right]
\tag{24}
\end{align*}
\end{figure*}

To determine the free parameters, in this study we propose the optimization in (21), where ${\bm{\Omega _\lambda}}$, ${\bm{\Omega _c}}$ and ${\bm{\Omega _\sigma}}$ denote the admissible sets of the parameter vectors $\bm{\lambda}$, $\bm{c}$ and $\bm{\sigma}$.

\emph{Remark}: It is worth noting that the objective function ${U_{\bm{\lambda ,c,\sigma}}}(T,Y)$ can be expressed as
\begin{equation}
\setcounter{equation}{22}
\label{eq22}
\begin{aligned}
{U_{\bm{\lambda ,c,\sigma}}}(T,Y) &= \frac{1}{2}QIP\left( e \right) \\
&- \frac{1}{2}{D_{ED}}\left( {\sum\limits_{i = 1}^m {{\lambda _i}{\kappa _{{\sigma _i}}}(\varepsilon  - {c_i})} \left\| {{p_e}(\varepsilon )} \right.} \right)
\end{aligned}
\end{equation}
where $QIP\left( e \right) = \int {{{\left( {{p_e}(\varepsilon )} \right)}^2}d\varepsilon }$ is the quadratic information potential (QIP) \cite{principe2010information} of the error $e$, and ${D_{ED}}\left( {.\left\| . \right.} \right)$ denotes the Euclidean distance between PDFs \cite{santamaria2002adaptive,heravi2018new}, defined by ${D_{ED}}\left( {p(x)\left\| {q(x)} \right.} \right) = \int {{{(p(x) - q(x))}^2}dx} $. Therefore, maximizing the objective function ${U_{\bm{\lambda ,c,\sigma}}}(T,Y)$ will try to maximize the QIP (or minimize Renyi's quadratic entropy) of the error and at the same time, minimize the Euclidean distance between the multi-Gaussian kernel function and the error's PDF.

If $N$ error samples $\left\{ {{e_j}} \right\}_{j = 1}^N$ are available, we have $\textbf{E}\left[ {\sum\limits_{i = 1}^m {{\lambda _i}{\kappa _{{\sigma _i}}}(e - {c_i})} } \right] \approx {\bm{\lambda} ^T}\bm{\tilde h}$, where $\bm{\tilde h} = \frac{1}{N}\sum\limits_{j = 1}^N {\bm{\tilde g}({e_j})}$, with $\bm{\tilde g}({e_j}) = {\left[ {{\kappa _{{\sigma _1}}}({e_j} - {c_1}), \cdots ,{\kappa _{{\sigma _m}}}({e_j} - {c_m})} \right]^T}$. Thus by (21), we have (23), where $\mathbf{\tilde K}$ is expressed in (24).

According to (23), the model $M$ and $3m$ free parameters are jointly optimized via maximizing the objective function ${\hat U_{\bm{\lambda ,c,\sigma}}}(T,Y) =  - \frac{1}{2}{\bm{\lambda} ^T}\mathbf{\tilde K}\bm{\lambda} {\rm{ + }}{\bm{\lambda} ^T}\bm{\tilde h}$. This is a very complicated optimization problem. To simplify the optimization, one can adopt an alternative optimization method: i) given a model (hence the $N$ error samples are given), we solve the free parameters by maximizing ${\hat U_{\bm{\lambda ,c,\sigma}}}(T,Y)$ (with error samples fixed); ii) after the free parameters are determined, we solve a new model by maximizing ${\hat U_{\bm{\lambda ,c,\sigma}}}(T,Y)$ (with free parameters fixed).
\begin{figure}
	\setlength{\belowcaptionskip}{0pt}
	\centering
    \includegraphics[height=2.6in]{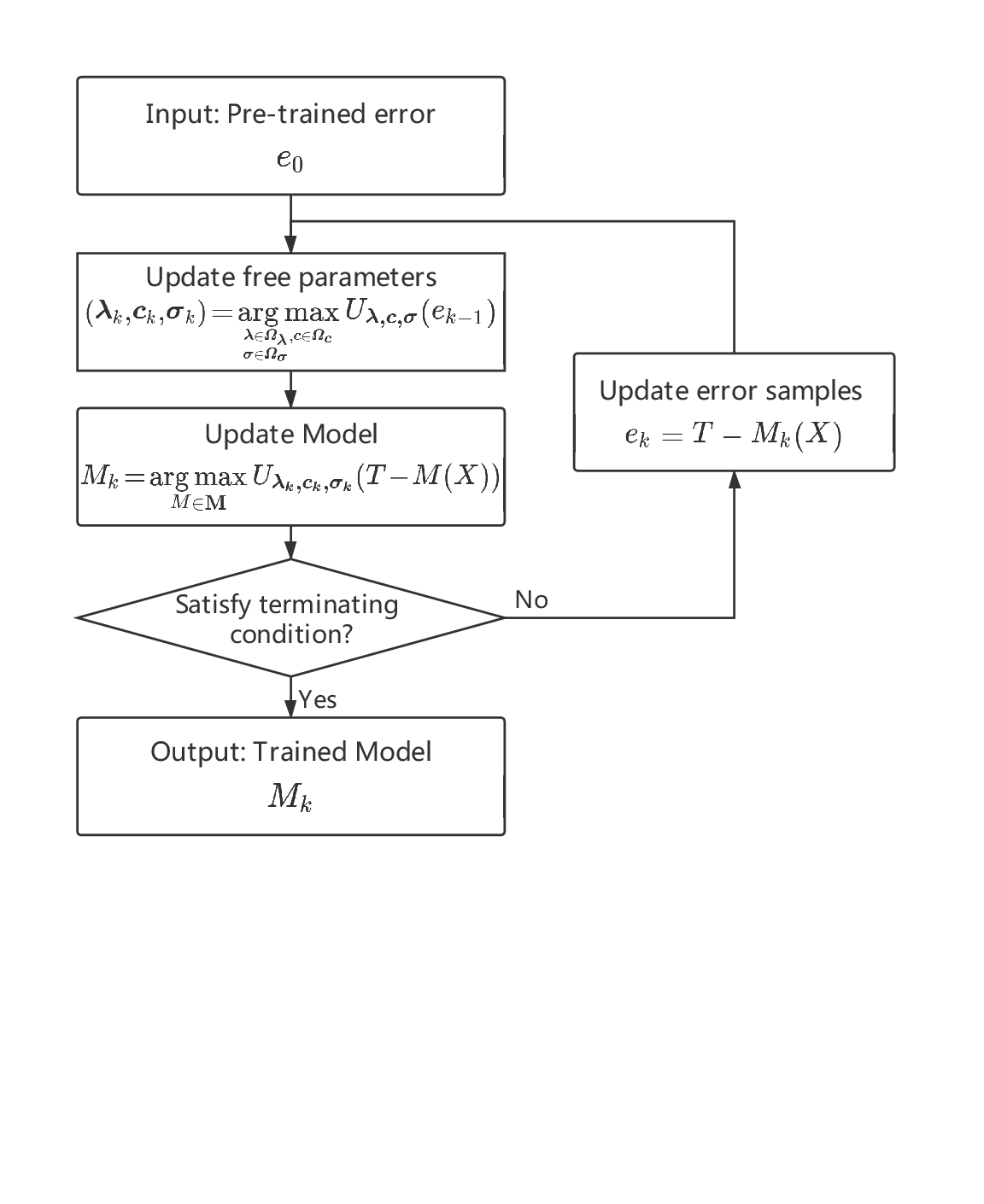}
	\caption{Alternative optimization for model and free parameters}
	\label{fig2}
\end{figure}

In a practical application, there are usually two approaches to find the free parameters and the optimal model. The first one is an online approach, in which the model is optimized by an iterative method and at each iteration, the $3m$ free parameters are determined based on the error samples at that iteration. The second one is a two-stage approach, which contains two stages: 1) train the model using a simple method (usually with very few free parameters) to obtain the error samples, and determine the $3m$ free parameters based on these errors; 2) train the model again under the MMKCC with the obtained free parameters, and during the training these free parameters are fixed. The above procedure can be repeated until convergence and the flow chart is shown in \reffig{fig2}.

\begin{figure*}[!t]
\setlength{\abovecaptionskip}{0pt}
\setlength{\belowcaptionskip}{0pt}
\centering
\subfigure[]{
\includegraphics[height=2in]{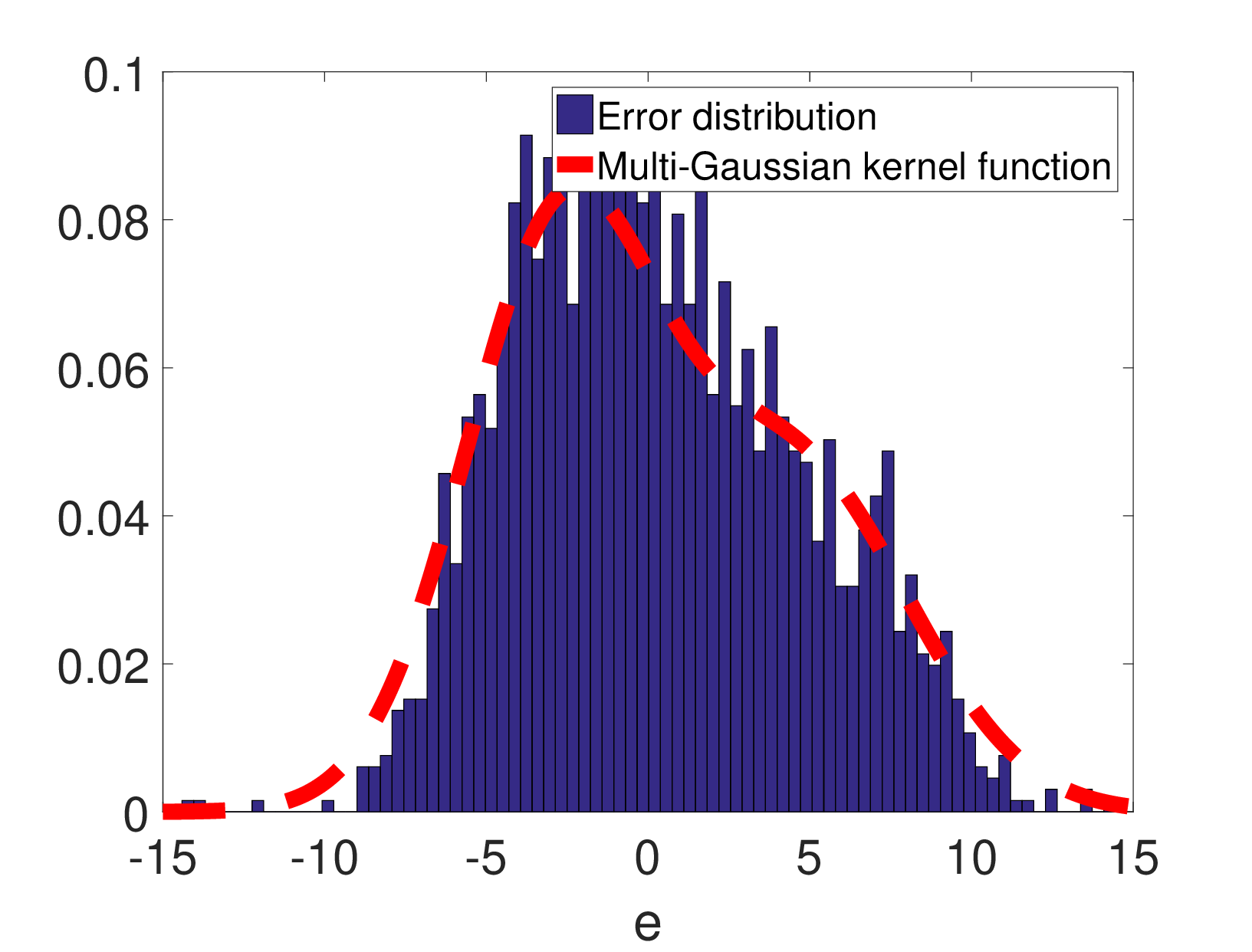}}
\subfigure[]{
\includegraphics[height=2in]{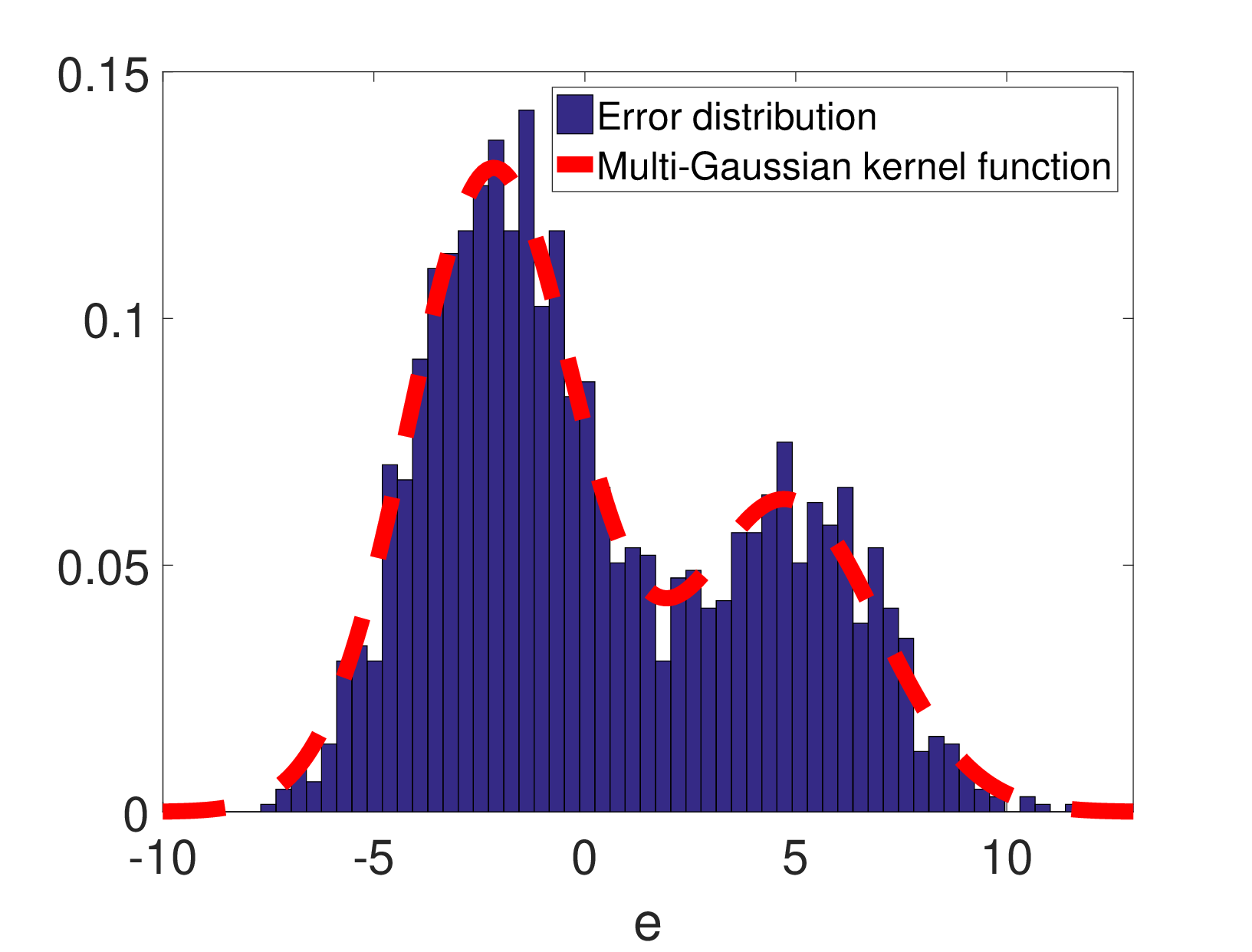}}
\subfigure[]{
\includegraphics[height=2in]{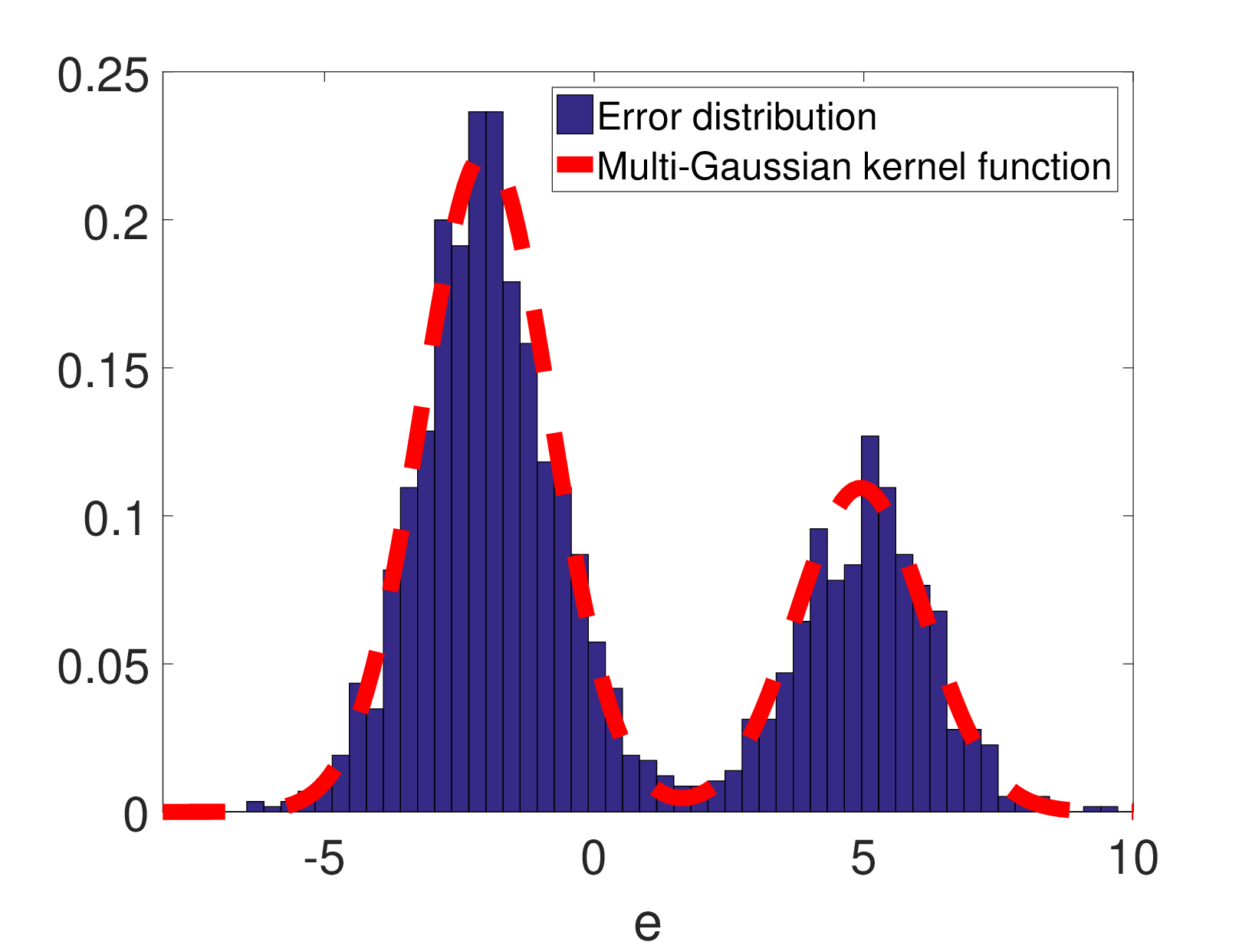}}
\subfigure[]{
\includegraphics[height=2in]{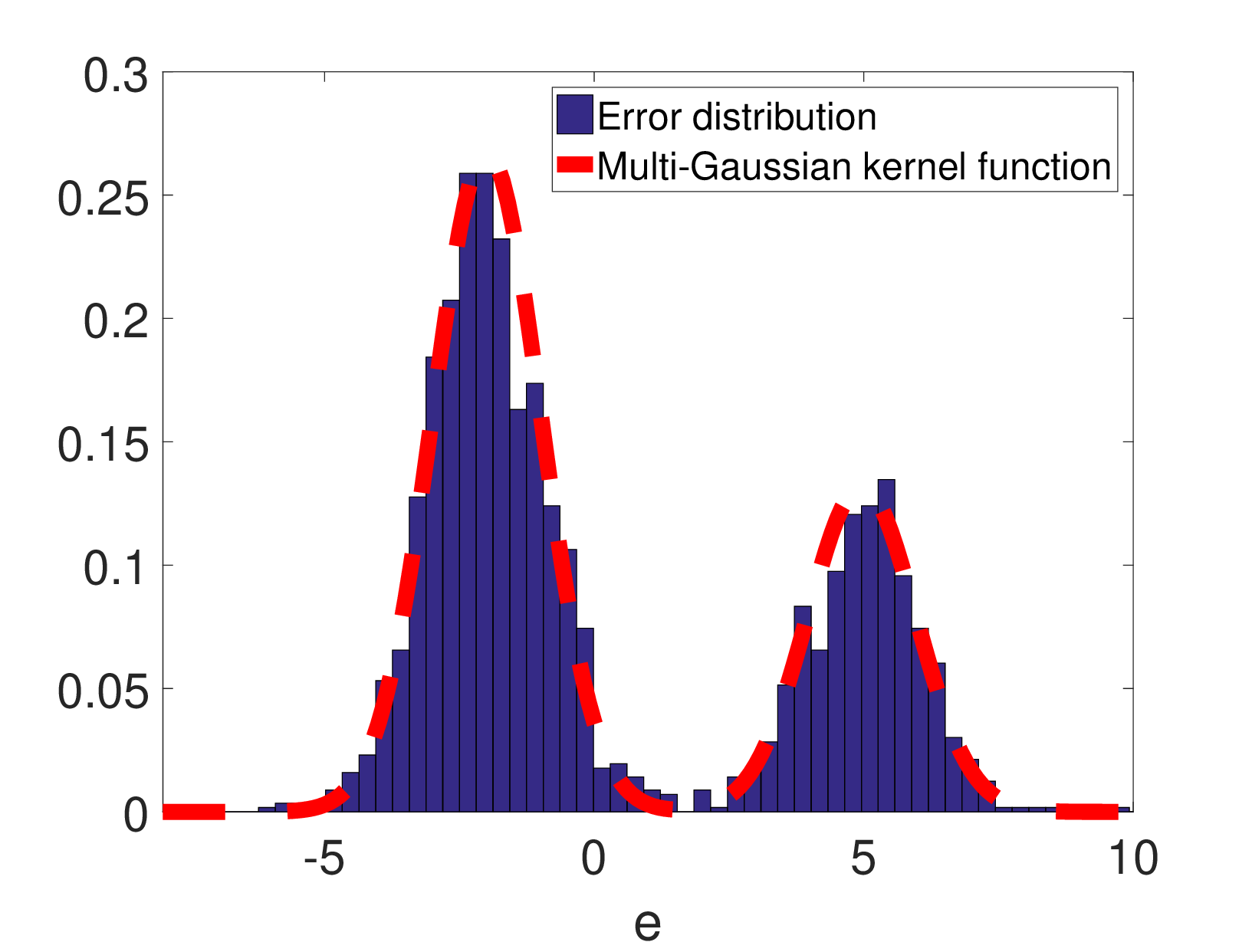}}
\caption{Error distributions and multi-Gaussian kernel functions at different fixed-point iterations: (a) first iteration; (b)second iteration, (c)third iteration, (d)fourth iteration}
\label{fig3}
\end{figure*}

Next, we describe how to determine the  $3m$ free parameters given a model. First, to simplify the optimization, we just apply some clustering technique such as the K-means on the error samples to obtain the center vector ${\bm{c}^*}$ (whose elements are the clustering centers). Then by (23), one can easily obtain the mixture coefficient vector:
\begin{equation}
\setcounter{equation}{25}
\label{eq25}
\begin{aligned}
{\bm{\lambda} ^{\rm{*}}} = {\bf{\tilde K}^{ - {\rm{1}}}}\bm{\tilde h}
\end{aligned}
\end{equation}
In order to avoid numerical problem in the matrix inversion, a regularized solution can be used:
\begin{equation}
\label{eq26}
\begin{aligned}
{\bm{\lambda} ^{\rm{*}}} = {(\bf{\tilde K} + \eta I)^{ - {\rm{1}}}}\bm{\tilde h}
\end{aligned}
\end{equation}
where $\eta$ is a regularization parameter. Substituting (26) into (23), we solve the bandwidth vector as follows:

\begin{equation}
\label{eq27}
\begin{aligned}
{\bm{\sigma} ^*} &= \mathop {\arg \max }\limits_{\bm{\sigma  \in {\Omega _\sigma }}}  - \frac{1}{2}{\left[ {{{(\mathbf{\tilde K} + \eta I)}^{ - {\rm{1}}}}\bm{\tilde h}} \right]^T}\mathbf{\tilde K}{(\mathbf{\tilde K} + \eta I)^{ - {\rm{1}}}}\bm{\tilde h} \\
&+ {\left[ {{{(\mathbf{\tilde K} + \eta I)}^{ - {\rm{1}}}}\bm{\tilde h}} \right]^T}\bm{\tilde h}
\end{aligned}
\end{equation}

In order to reduce the computational complexity of the optimization problem in (27), one can alternately optimize every dimension of the bandwidth vector over a finite set of values. Specifically, given a finite set of bandwidths ${{\bm{\Omega} _\sigma}}$, we optimize each element of the bandwidth vector $\bm{\sigma}$ one by one and repeat this procedure until convergence. The proposed procedure for free parameters determination is summarized in \textbf{Algorithm 2}.

\begin{algorithm}
	\renewcommand{\algorithmicrequire}{\textbf{Input:}}
	\renewcommand{\algorithmicensure}{\textbf{Output:}}
	\caption{Determination of the free parameters}
	\label{alg:1}
	\begin{algorithmic}[1]
		\REQUIRE error samples $\left\{ {{e_j}} \right\}_{j = 1}^N$, parameter dimension $m$, regularization parameter $\eta$, a finite set of bandwidths ${{\bm{\Omega} _\sigma}}$ and  initialize ${\sigma _1} =  \cdots  = {\sigma _m} = {\sigma _0}$.
		\ENSURE  free parameters $\bm{{\lambda ^*},{c^*},{\sigma ^*}}$ \\
        \STATE Determine the center vector $\bm{c^*}$ by applying the K-means clustering on the error samples $\left\{ {{e_j}} \right\}_{j = 1}^N$
        \STATE Alternately optimize every dimension of the bandwidth vector $\bm{\sigma}$ and repeat $S$ times:
        \FORALL{$s = 1,2,...,S$}
        \FORALL{$i = 1,2,...,m$}
        \STATE ${\sigma_i^*} = \mathop {\arg \max }\limits_{\sigma_i \in {{\bm{\Omega} _\sigma }}}  - \frac{1}{2}{\left[ {{{(\mathbf{\tilde K} + \eta I)}^{ - {\rm{1}}}}\bm{\tilde h}} \right]^T}\mathbf{\tilde K}{(\mathbf{\tilde K} + \eta I)^{ - {\rm{1}}}}\bm{\tilde h} + {\left[ {{{(\mathbf{\tilde K} + \eta I)}^{ - {\rm{1}}}}\bm{\tilde h}} \right]^T}\bm{\tilde h}$, with $\bm{c=c^*}$ and other $m-1$ elements of $\bm{\sigma}$ being fixed
        \ENDFOR
        \ENDFOR
		\STATE Compute ${\bm{\lambda} ^{\rm{*}}} = {(\bf{\tilde K} + \eta I)^{ - {\rm{1}}}}\bm{\tilde h}$ with $\bm{\sigma=\sigma^*}$ and $\bm{c=c^*}$
\textbf{Return:} $\bm{\lambda^*,\sigma^*,c^*}$
	\end{algorithmic}
\end{algorithm}

\begin{table*}
\centering
\caption{RMSEs and computing times(sec) of different learning criteria}\label{RMSE}
\begin{tabular}{ccccccc}
  \hline
   & &MSE&MCC&MMCC&MMKCC \\
  \hline
  \multirow{2}{*}{case 1)}&RMSE& $0.5427\pm0.3175$  &0.0881$\pm0.0431$  &$0.0831\pm0.0375$ &$\textbf{0.0342}\pm\textbf{0.0259}$\\
  &TIME(sec)&N/A&$0.0832\pm0.0020$ &$0.1027\pm0.0026$&$0.3328\pm0.0070$\\
  \multirow{2}{*}{case 2)}&RMSE& $0.5031\pm0.2483$  &$0.0754\pm0.0414$  &$0.0674\pm0.0334$ &$\textbf{0.0224}\pm\textbf{0.0115}$\\
  &TIME(sec)&N/A&$0.0814\pm0.0018$ &$0.1014\pm0.0024$&$0.3415\pm0.0075$\\
  \multirow{2}{*}{case 3)}&RMSE& $0.5494\pm0.3418$  &$0.0391\pm0.0191$  &$0.0353\pm0.0176$ &$\textbf{0.0335}\pm\textbf{0.0168}$\\
  &TIME(sec)&N/A&$0.0841\pm0.0022$ &$0.1021\pm0.0027$&$0.3297\pm0.0068$\\
  \hline
\end{tabular}
\end{table*}

\section{EXPERIMENTAL RESULTS}

In this section, we present experimental results to demonstrate the desirable performance of learning methods under the proposed MMKCC criterion (i.e. the FP-MMKCC algorithm). Without explicit mention, the dimension number is $m=2$, the regularization parameter is $\eta  = 10^{-4}$ and the iteration number $S$ is $S=3$.

\subsection{Linear Regression}
First, we consider a simple linear regression example where the input-target samples are generated by a two-dimensional linear system:  ${t_i} = {\bm{\beta}^*}^T{\bm{x}_i} + {\rho _i}$, where ${\bm{\beta}^*} = {[1,2]^T}$ is the weight vector to be estimated, and ${\rho _i}$ denotes an additive noise. The input samples $\{ {\bm{x}_i}\}$ are uniformly distributed over $[ - 2.0,2.0] \times [ - 2.0,2.0]$. The noise ${\rho _i}$ comprises two mutually independent noises, namely the inner noise ${B _i}$ and the outlier noise ${O_i}$. Specifically, ${\rho _i}$ is given by $\rho {}_i = {\rm{(}}1 - {g_i}{\rm{)}}{B_i} + {g_i}{O_i}$, where ${g _i}$ is a binary variable with probability mass $\Pr {\rm{\{ }}{g_i} = 1{\rm{\} }} = p$, $\Pr {\rm{\{ }}{g_i} = 0{\rm{\} }} = 1 - p$, $(0 \le p \le 1)$, which is assumed to be independent of both $B_i$ and $O_i$. In this example, $p$ is set at $0.1$, and the outlier $O_i$ is drawn from a zero-mean Gaussian distribution with variance $10000$. As for the inner noise $B_i$, we consider three cases: 1) $B_i \sim$ 0.5$\mathcal{N}$(4.0,1.0)+0.5$\mathcal{N}$(-4.0,1.0), where $\mathcal{N}(u,\sigma^2)$ denotes the Gaussian density function with mean $u$ and variance $\sigma^2$ ; 2) $B_i \sim$ 1/3$\mathcal{N}$(5.0,1.0)+2/3$\mathcal{N}$(-2.0,1.0). 3) $B_i \sim$  0.5$\mathcal{N}$(0,1.0)+0.5$\mathcal{N}$(0,5.0). The root mean squared error (RMSE) is employed to measure the performance, computed by $RMSE = \sqrt {\frac{1}{2}{{\left\| {{\bm{\beta}_k} - {\bm{\beta}^*}} \right\|}^2}}$, where $\bm{\beta}_k$ and $\bm{\beta}^*$ denote the estimated and the target weight vectors respectively.

\begin{figure}
	\setlength{\abovecaptionskip}{0pt}
	\setlength{\belowcaptionskip}{0pt}
	\centering
    \includegraphics[height=1.6in]{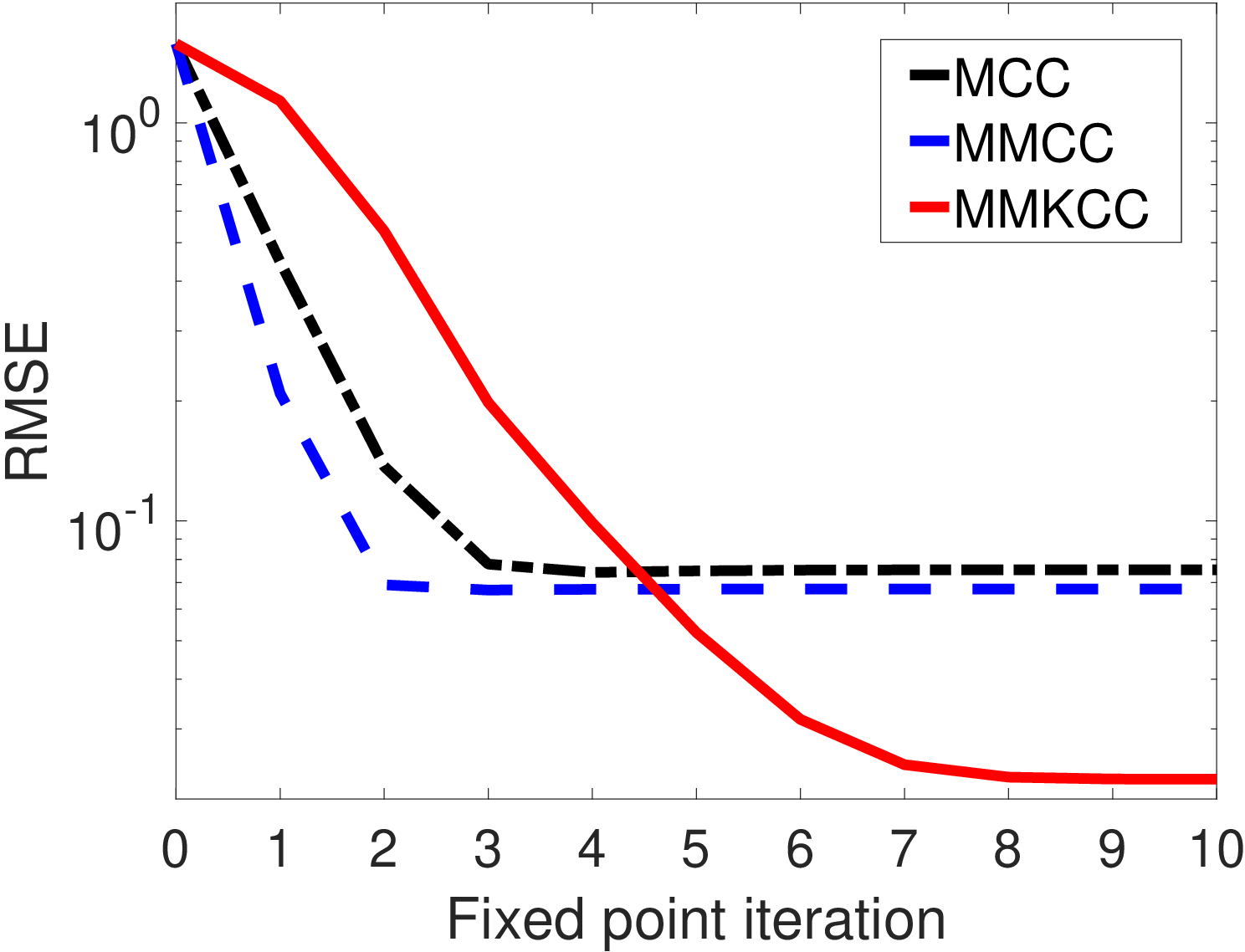}
	\caption{RMSE convergence curves of different learning criteria}
	\label{fig4}
\end{figure}

We compare the performance of four learning criteria, namely MSE, MCC, MMCC and MMKCC. For MSE, there is a closed-form solution, so no iteration is needed. For MCC, MMCC and MMKCC, a fixed-point iteration is used to solve the model (see the \textbf{Algorithm 1} for the fixed-point algorithm under MMKCC). The mean $\pm$ deviation results of the RMSEs and computing times over 100 Monte Carlo runs are presented in Table II. In the simulation, the sample number is $N$ = 400, the fixed-point iteration number is $K$ = 10, and the initial weight vector is set to ${\beta _0} = {[{\rm{0}},{\rm{0}}]^T}$. For each learning criterion, the parameters are experimentally selected to achieve the best results, except that the free parameters of MMKCC are determined by \textbf{Algorithm 2}. The finite set ${\bm{\Omega} _\sigma}$ in \textbf{Algorithm 2} is equally spaced over [0.1, 2.0] with step size 0.2. The simulations are carried out with MATLAB 8.6 running in Core 4 Quad, 3.4-GHZ CPU with 20-GB RAM. From Table II, we observe: i) MCC, MMCC and MMKCC can significantly outperform MSE although they have no closed-form solution; ii) MMKCC can achieve better performance than MCC and MMCC especially for noises with multi-peak or asymmetric distributions; iii) although the MMKCC is computationally more expensive than MCC and MMCC, the computing times of three learning criteria are in the same order of magnitude. For the noise case 2), the error distributions and multi-Gaussian kernel functions (determined by \textbf{Algorithm 2} after each fixed-point iteration) at different fixed-point iterations under MMKCC are shown in \reffig{fig3}. As expected, the multi-Gaussian kernel function matches the error distribution very well at every iteration (as discussed earlier, the free parameters in MMKCC have been optimized to minimize the Euclidean distance between the multi-Gaussian kernel function and the error PDF). The average RMSE convergence curves of three learning criteria are illustrated in \reffig{fig4}.

\begin{table}
	\renewcommand\arraystretch{1.5}
	\setlength{\abovecaptionskip}{0pt}
	\setlength{\belowcaptionskip}{0pt}
	\centering
	\caption{Specification of the datasets}
	\begin{tabular}{cccc}
		\toprule
		\multirow{2}*{Datasets} & \multirow{2}*{Features} & \multicolumn{2}{c}{Observations}\\
		\cline{3-4}
		&{}&Training&Testing\\
		\midrule
		Servo	&5	&83	&83\\
		Concrete	&9	&515 &515\\
        Airfoil	&5	&751	&751\\
        Yacht	&6	&154	&154\\
		\bottomrule
	\end{tabular}
\end{table}

\begin{table*}[]\small
	\renewcommand\arraystretch{1.5}
	\setlength{\abovecaptionskip}{0pt}
	\setlength{\belowcaptionskip}{5pt}
	\centering
	\caption{RMSEs and training times(sec) of several RSCN algorithms}
	\begin{tabular}{ccccccccc}
		\toprule
		\multicolumn{1}{c}{\multirow{2}*{Datasets}} & \multicolumn{2}{c}{{RSCN}}& \multicolumn{2}{c}{{RSC-MCC}}& \multicolumn{2}{c}{{RSC-MMCC}}& \multicolumn{2}{c}{{RSC-MMKCC}}\\
		\cline{2-9}
		&\multicolumn{1}{c}{{RMSE}} &\multicolumn{1}{c}{{Training Time}}&\multicolumn{1}{c}{{RMSE}} &\multicolumn{1}{c}{Training Time} &\multicolumn{1}{c}{RMSE} &\multicolumn{1}{c}{Training Time} &\multicolumn{1}{c}{RMSE} &\multicolumn{1}{c}{Training Time}\\
        Hardware
        & \makecell[c]{$0.1000\pm$\\$0.0325{\kern8pt}$} & \makecell[c]{$0.0129\pm$\\$0.0021{\kern8pt}$} & \makecell[c]{$0.0754\pm$\\$0.0254{\kern8pt}$}
        & \makecell[c]{$0.1074\pm$\\$0.0256{\kern8pt}$} & \makecell[c]{$0.0734\pm$\\$0.0223{\kern8pt}$} & \makecell[c]{$0.1024\pm$\\$0.0199{\kern8pt}$}
        & \makecell[c]{$\textbf{0.0719}\pm$\\$\textbf{0.0208}{\kern8pt}$} & \makecell[c]{$0.3464\pm$\\$0.0389{\kern8pt}$}\\
        Servo
        & \makecell[c]{$0.1293\pm$\\$0.0322{\kern8pt}$} & \makecell[c]{$0.0237\pm$\\$0.0020{\kern8pt}$} & \makecell[c]{$0.1211\pm$\\$0.0237{\kern8pt}$}
        & \makecell[c]{$0.0462\pm$\\$0.0098{\kern8pt}$} & \makecell[c]{$0.1181\pm$\\$0.0228{\kern8pt}$} & \makecell[c]{$0.0591\pm$\\$0.0139{\kern8pt}$}
        & \makecell[c]{$\textbf{0.1169}\pm$\\$\textbf{0.0224}{\kern8pt}$} & \makecell[c]{$0.2896\pm$\\$0.0188{\kern8pt}$}\\
        Yacht
        & \makecell[c]{$0.0484\pm$\\$0.0118{\kern8pt}$} & \makecell[c]{$0.1458\pm$\\$0.0089{\kern8pt}$} & \makecell[c]{$0.0427\pm$\\$0.0149{\kern8pt}$}
        & \makecell[c]{$0.3273\pm$\\$0.0422{\kern8pt}$} & \makecell[c]{$0.0400\pm$\\$0.0139{\kern8pt}$} & \makecell[c]{$0.3348\pm$\\$0.0389{\kern8pt}$}
        & \makecell[c]{$\textbf{0.0385}\pm$\\$\textbf{0.0166}{\kern8pt}$} & \makecell[c]{$0.8753\pm$\\$0.0456{\kern8pt}$}\\
        Airfoil
        & \makecell[c]{$0.0923\pm$\\$0.0057{\kern8pt}$} & \makecell[c]{$1.7234\pm$\\$0.0811{\kern8pt}$} & \makecell[c]{$0.0905\pm$\\$0.0065{\kern8pt}$}
        & \makecell[c]{$3.0372\pm$\\$0.1145{\kern8pt}$} & \makecell[c]{$0.0900\pm$\\$0.0056{\kern8pt}$} & \makecell[c]{$3.0522\pm$\\$0.1230{\kern8pt}$}
        & \makecell[c]{$\textbf{0.0893}\pm$\\$\textbf{0.0055}{\kern8pt}$} & \makecell[c]{$3.9181\pm$\\$0.1379{\kern8pt}$}\\
		\bottomrule
	\end{tabular}
\end{table*}

\begin{table*}[]
\centering
\caption{Testing RMSEs of TDNNs trained under different criteria}\label{RMSE}
\begin{tabular}{cccccc}
  \hline
   &MSE&MCC&MMCC&MMKCC \\
  \hline
  RMSE& 0.0427&0.0309&0.0302&\textbf{0.0277}\\
  \hline
\end{tabular}
\end{table*}

\subsection{Non-linear regression with benchmark datasets}

In the second example, we show the superior performance of the MMKCC criterion in nonlinear regression with five benchmark data sets from UCI machine learning repository \cite{frank2010uci}. The descriptions of the data sets are given in Table III. In the experiment, the training and testing samples from each data set are randomly chosen and the data values are normalized into [0, 1.0]. The robust stochastic configuration networks (RSCN) is adopted as the regression model to be trained, which is a linear-in-parameter (LIP) model with randomly generated hidden nodes \cite{wang2017stochastic,wang2017robust,li2019robust}. Under the MMKCC, the model is trained by the fixed-point iterative algorithm in \textbf{Algorithm 1} and we call it the RSC-MMKCC algorithm. In this example, the performance of the RSC-MMKCC is compared with that of several other stochastic configuration networks (SCN) based algorithms, including RSCN\cite{wang2017robust}, RSC-MCC\cite{li2019robust} and RSC-MMCC, where the RSC-MMCC can be viewed as RSC-MMKCC with $\bm{c}=\mathbf{0}$. The parameters of each algorithm are selected through fivefold cross-validation, except that the free parameters of MMKCC are determined by \textbf{Algorithm 2}. The finite set ${\bm{\Omega} _\sigma}$ in \textbf{Algorithm 2} is equally spaced over [0.1, 3.0] with step size 0.1. The training and testing RMSEs over 100 runs are presented in Table IV. Clearly, the RSC-MMKCC outperforms the RSCN, RSC-MCC and RSC-MMCC for all the data sets.

\emph{Remark}: The parameter setting method of \textbf{Algorithm 2} may have similar or worse performances compared with the cross-validation. However, the cross-validation will take a lot of time when the parameter space is very large ($3m$ parameters for MKC). Thus, the cross-validation approach for RSC-MMKCC is not practical. Actually, the proposed parameter setting method is computationally much simpler than the cross-validation but can still achieve desirable performances (RSC-MMKCC performs better than RSC-MMCC and RSC-MCC in this example).

\subsection{Chaotic time series prediction}
\begin{figure}[t]
	\setlength{\abovecaptionskip}{0pt}
	\setlength{\belowcaptionskip}{0pt}
	\centering
    \includegraphics[height=1.7in]{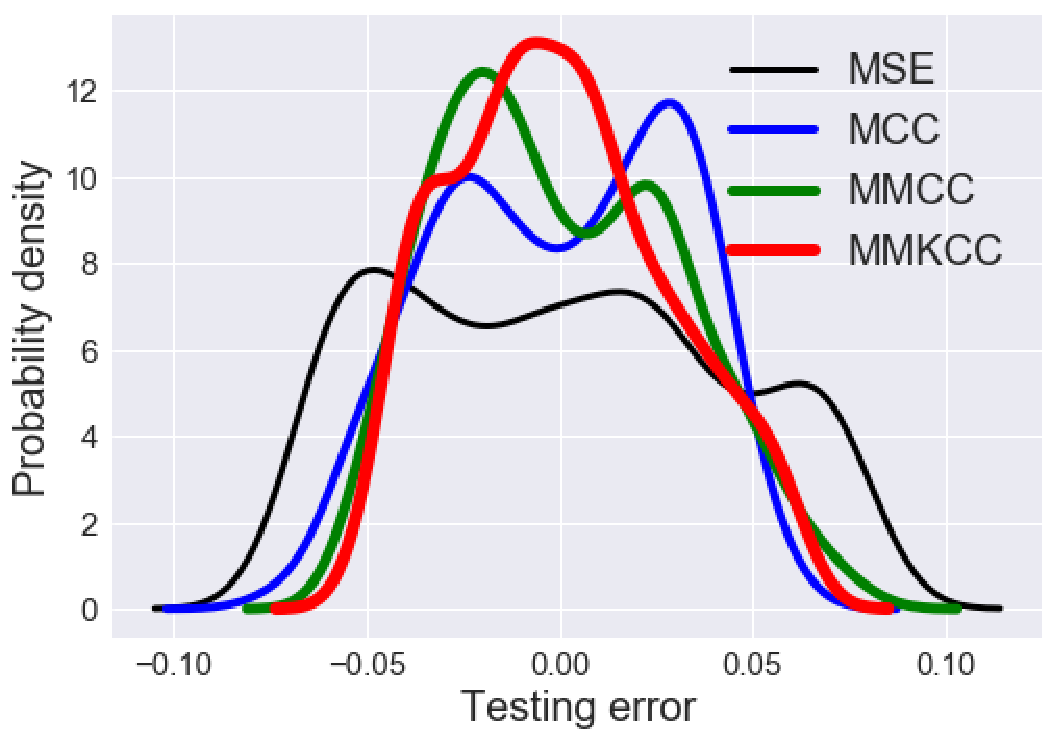}
	\caption{Testing error PDFs of TDNNs trained under different criteria}
	\label{fig5}
\end{figure}

In the third example, we apply different learning criteria (MSE, MCC, MMCC, MMKCC) to train a time delay neural network (TDNN)\cite{waibel1995phoneme} to predict the Mackey-Glass chaotic time series\cite{kuo1993nonlinear}. The TDNN has a single hidden layer and six nonlinear processing elements in the hidden layer, and its inputs consist of six delayed values. A sigmoid nonlinearity was used in each of the hidden processing elements, while the output processing element was linear. The sequence for training has an additive noise, and the training samples are generated by
\begin{equation}\small
\label{eq26}
\begin{aligned}
x(t) =  - bx(t - 1) + \frac{{ax(t - \tau )}}{{1 + x{{(t - \tau )}^{10}}}} + {\rho _t}
\end{aligned}
\end{equation}
with $b=0.1$, $a=0.2$,$\tau {\rm{ = 30}}$ and ${\rho _t}\sim 0.45\mathcal{N}(-0.05,0.05) + 0.45\mathcal{N}(0.05,0.05)+0.1\mathcal{N}(0,0.2)$. The TDNN is trained to predict the next sample of the time series by using six previous samples, with a segment of 200 samples. The trained networks are tested on clean data set (without additive noise) of length 1000. The kernel size of MCC is experimentally set at $\sigma=2.0$, and the kernel sizes of MMCC are ${\sigma _1} = {\rm{1}}.0$, ${\sigma _2} = {\rm{2}}.0$, and the mixture coefficient in MMCC is $\alpha  = 0.8$. For MMKCC, the finite set ${\bm{\Omega} _\sigma}$ is equally spaced over [0.1, 3.0] with step size 0.1. The PDFs of the testing error averaged over 10 Monte Carlo runs are illustrated in \reffig{fig5} and the corresponding testing RMSEs are presented in Table V. Evidently, the TDNN trained under MMKCC achieves the best performance with the most concentrated error distribution and the lowest RMSE.

\begin{table*}[]
	\setlength{\abovecaptionskip}{0pt}
	\setlength{\belowcaptionskip}{5pt}
	\centering
	\caption{Classification Accuracies of Different Algorithms on the Data Set IIa of BCI Competition IV}
	\begin{tabular}{ccccccccc}
		\toprule
		Subject & KNN & SVM & RSCN & RSC-MCC & RSC-MMCC & RSC-MMKCC\\
        \midrule
        A01
        & $70.83$ & $72.92$ & $73.97\!\pm\!1.87$ & $74.20\!\pm\!1.79$ & $74.93\!\pm\!2.03$ & $\textbf{75.69}\!\pm\!\textbf{1.89}$\\
        A02
        & $43.40$ & $46.88$ & $46.32\!\pm\!1.46$ & $46.25\!\pm\!1.57$ & $46.47\!\pm\!1.79$ & $\textbf{46.88}\!\pm\!\textbf{1.41}$\\
        A03
        & $74.65$ & $76.39$ & $76.71\!\pm\!1.28$ & $76.92\!\pm\!1.30$ & $76.77\!\pm\!1.33$ & $\textbf{76.80}\!\pm\!\textbf{1.30}$\\
        A04
        & $55.21$ & $62.15$ & $60.72\!\pm\!1.85$ & $60.51\!\pm\!1.84$ & $60.59\!\pm\!1.69$ & $\textbf{61.46}\!\pm\!\textbf{1.65}$\\
        A05
        & $35.07$ & $35.07$ & $\textbf{40.56}\!\pm\!\textbf{1.68}$ & $40.41\!\pm\!2.10$ & $40.32\!\pm\!1.92$ & $40.28\!\pm\!1.90$\\
        A06
        & $40.63$ & $42.01$ & $44.41\!\pm\!2.02$ & $\textbf{44.50}\!\pm\!\textbf{2.03}$ & $43.90\!\pm\!1.77$ & $43.75\!\pm\!1.79$\\
        A07
        & $75.35$ & $77.78$ & $78.61\!\pm\!1.22$ & $80.03\!\pm\!1.68$ & $79.36\!\pm\!1.65$ & $\textbf{80.56}\!\pm\!\textbf{1.52}$\\
        A08
        & $73.96$ & $79.51$ & $79.22\!\pm\!1.45$ & $79.76\!\pm\!2.04$ & $\textbf{81.18}\!\pm\!\textbf{1.32}$ & $79.51\!\pm\!1.29$\\
        A09
        & $81.25$ & $79.17$ & $79.63\!\pm\!2.48$ & $80.28\!\pm\!2.45$ & $79.94\!\pm\!2.35$ & $\textbf{86.11}\!\pm\!\textbf{2.35}$\\
        Mean
        & $61.15$ & $63.54$ & $64.46\!\pm\!1.70$ & $64.76\!\pm\!1.87$ & $64.83\!\pm\!1.76$ & $\textbf{65.67}\!\pm\!\textbf{1.68}$\\
        \bottomrule
	\end{tabular}
\end{table*}

\begin{table*}[]
	\setlength{\abovecaptionskip}{0pt}
	\setlength{\belowcaptionskip}{5pt}
	\centering
	\caption{The $p$-value of the Paired Sample T-Test between Classification Accuracies of MMCC and MMKCC}
	\begin{tabular}{cccccccccccccccccccccc}
		\toprule
		Subject & A01 & A02 & A03 & A04 & A05 &A06     & A07 & A08 & A09 & Mean Accuracy\\
        \midrule
        $p$-value & $0.0433$ & $0.2206$ & $0.8867$ & $0.0067$ & $0.8954$ & $0.6432$ & $0.0002$ & $4.0*10^{-09}$ & $8.7*10^{-21}$ & $7.4*10^{-08}$\\
        \bottomrule
	\end{tabular}
\end{table*}

\subsection{EEG Classification}
Electroencephalography (EEG) is a kind of multichannel electrophysiological signal recorded by electrodes placed on the scalp typically (also subdurally or in the cerebral cortex), which plays an important role in the brain–computer interface (BCI) systems \cite{lotte2018review, he2020brain}. A BCI system can be defined as a system that translates the brain activity patterns into commands for an interactive application \cite{lotte2018review}, and for an EEG-based BCI system, the goal is to effectively recognize the brain patterns of a user from the collected EEG signals. In view of the fact that EEG recordings are often contaminated by various artifacts, such as artifacts due to electrode displacement, motion artifacts, ocular artifacts and so on \cite{islam2016methods}, the proposed RSC-MMKCC could be a good candidate for the EEG classifier due to its excellent flexibility and robustness.

The benchmark data set adopted here is the data set IIa of BCI Competition IV \cite{tangermann2012review}, which consists of EEG data from 9 subjects. The BCI paradigm consisted of four different motor imagery tasks, i.e., left hand, right hand, both feet and tongue. Each subject includes two sessions, and each session is comprised of 6 runs. One run consists of 48 trials (12 for each class), yielding a total of 288 trials per session.

Considering the EEG feature extraction, the common spatial pattern (CSP) is an effective approach for multichannel EEG data concerning motor imagery tasks \cite{lotte2010regularizing}. We adopt the CSP combined with the one-versus-one (OVO) \cite{rocha2013multiclass} approach, which transforms the four-class classification problem into six cases of two-class classification. The first two spatial filters that correspond to the largest objective function values are used, and vice versa. Then the log variances of the spatially ﬁltered EEG signals are used as the input features for classiﬁers. As a result, each trial is assigned with a 24-D feature.

Besides the aforementioned RSCN, RSC-MCC, RSC-MMCC and RSC-MMKCC, k-nearest neighbor (KNN) \cite{sun2010adaptive} and support vector machine (SVM) \cite{suykens1999least} are also chosen as the classifiers in EEG classification tasks for comparison. The parameters of each algorithm are selected through fivefold cross-validation, except that the free parameters of MMKCC are determined by \textbf{Algorithm 2}. Table VI shows the averaged classification accuracies of different algorithms after 30 Monte Carlo runs, and the highest accuracy for each subject is marked in bold. One can observe that the proposed RSC-MMKCC can achieve a higher classification accuracy on most subjects, and it also has the highest average classification accuracy over all the subjects. The standard deviations of KNN and SVM are zero since no random projection mechanism is adopted in them. With the Null Hypothesis $H_0:\mu_d=0$, where $d$ is the difference between the accuracies on pair, the $p$-values of the paired sample t-test between the classification accuracy of MMKCC and MMCC are shown in Table VII. With the significance level $\alpha=0.05$, we reject $H_0$ and state that we have significant evidence that accuracy difference between MMKCC and MMCC is NOT 0 for subject A01, A04, A07, A08 A09 and the mean accuracy, while the others are insignificant. Thus we can conclude that the MMKCC generally has better performance than the MMCC in statistics

\section{Conclusion}
A new generalized version of correntropy, called multi-kernel correntropy (MKC), was proposed in this study, where the kernel function is a mixture Gaussian kernel with different widths and centers. The original correntropy and the recently proposed mixture correntropy are both special cases of the new definition. Some important properties of the MKC were presented. In addition, a novel approach was proposed to determine the free parameters of MKC when used in supervised learning. The superior performance of the proposed learning method has been confirmed by experimental results of linear regression, nonlinear regression with benchmark datasets, chaotic time series prediction and EEG classfication.

\begin{appendices}
\renewcommand{\theequation}{\thesection.\arabic{equation}}
\setcounter{equation}{0}

\section{Convergence Analysis of FP-MMKCC}
The contraction mapping theorem (also known as the \emph{Banach Fixed-Point Theorem}) provides an effective way to prove the convergence of a fixed-point algorithm \cite{zhang2015convergence, chen2015convergence}.

The FP-MMKCC alogrithm can be described as
\begin{equation}
\begin{aligned}
  \bm{\beta}_{k}&=({\bf{H}^T}\bm{\Lambda}\bf{H} + \gamma '\mathbf{I})^{-1} ({\bf{H}^T}\bm{\Lambda}\bf{T} - {\bf{H}^T}\bm{\theta})\\
       &=(\mathbf{A}(\bm{\beta}_{k-1})+\gamma'\mathbf{I})^{-1}\mathbf{B}(\bm{\beta}_{k-1})\\
       &=f(\bm{\beta}_{k-1}).
\end{aligned}
\end{equation}
According to the contraction mapping theorem, the convergence of a fixed-point algorithm is guaranteed if $\exists \delta >0$ and $0<\alpha<1$ such that if the initial weight vector $\big\|\bm{\beta}_0\big\|_p \leq \delta$, and $\forall \bm{\beta} \in \{\bm{\beta}\in\mathbb{R}^L : \big\|\bm{\beta}_0\big\|_p \leq \delta\}$, it holds that
\begin{equation}\small
  \left\{
  \begin{array}{c}
    \big\|\bm{f}(\bm{\beta})\big\|_p \leq \delta\\
    \big\|\nabla_{\bm{\beta}}\bm{f}(\bm{\beta})\big\|_p = \bigg\|\dfrac{\partial \bm{f}(\bm{\beta})}{\partial \bm{\beta}^T}\bigg\|_p \leq \alpha,
  \end{array}
  \right.
\end{equation}
in which ``$\|\cdot\|_p$'' denotes an $\ell_p$-norm for a vector or an induced norm of a matrix, defined by $\big\|A\big\|_p=\underset{\|X\|_p\neq0}{\max}\big\|AX\big\|_p/\big\|X\big\|_p$, with $p\geq1, A\in\mathbb{R}^{m\times m}, X\in\mathbb{R}^{m\times 1}$, and $\nabla_{\bm{\beta}}\bm{f}(\bm{\beta})$ denotes the $m\times m$ Jacobian matrix of $\bm{f}(\bm{\beta})$ with respect to $\bm{\beta}$, given by \vspace{-1ex}
\begin{equation}\label{nablaf}\small
  \nabla_{\bm{\beta}}\bm{f}(\bm{\beta})\!=\!\begin{bmatrix}
                            \dfrac{\partial}{\partial \beta_1}\bm{f}(\bm{\beta}) & \dfrac{\partial}{\partial \beta_2}\bm{f}(\bm{\beta}) & \cdots & \dfrac{\partial}{\partial \beta_m}\bm{f}(\bm{\beta})
                          \end{bmatrix},
\end{equation}
in which $\beta_s$ is the $s$-th variable of $\bm{\beta}$.

To obtain a sufficient condition to guarantee the convergence of the FP-MMKCC algorithm, we put forward two theorems below. For simplicity, we denote kernel bandwidths as $\sigma_i=\mu_i\sigma$, where $\mu_i$ is a positive constant. \vspace{-1ex}

\begin{theorem}
  If $\delta>\xi=\dfrac{\vartheta\sqrt{m}}{\lambda_{\min}(\sum\limits_{j=1}^{N} \sum\limits_{i=1}^{m} \dfrac{\lambda_i}{\mu_i}\bm{h}_i{\bm{h}_i}^T)+\lambda_r}$ and $\sigma\geq\sigma^*$, where $\vartheta=\sum\limits_{j=1}^{N} \sum\limits_{i=1}^{m} \dfrac{\lambda_i}{\mu_i^3}\big|t_j\!-\!c_i\big| \big\|\bm{h}_j\big\|_1$, $\lambda_{\min}[\cdot]$ denotes the minimum eigenvalue of the matrix term and $\sigma^*$ is the solution of equation $\varphi(\sigma)=\dfrac{\vartheta\sqrt{m}}{\lambda_{\min}(\theta)+\lambda_r}=\delta,\sigma\in(0,\infty)$ with $\theta=\sum\limits_{j=1}^{N} \sum\limits_{i=1}^{m} \dfrac{\lambda_i}{\mu_i^3} \exp{\Big(-\dfrac{(\delta\|\bm{h}_j\|_1+|t_j-c_i|)^2}{2\mu_i^2\sigma^2}\Big)} \bm{h}_j^T\bm{h}_j$. Then $\big\|\bm{f}(\bm{\beta})\big\|_1 \leq \delta$ for all $\bm{\beta} \in \{\bm{\beta} \in \mathbb{R}^L : \big\|\bm{\beta}\big\|_1 \leq \delta\}$.
\end{theorem}
\begin{IEEEproof}
  The induced matrix norm is compatible with the corresponding vector $\ell_p$-norm, hence
  \begin{equation}\label{proof1.1}
  \begin{aligned}
      \big\|\bm{f}(\bm{\beta})\big\|_1 &= \big\|[\mathbf{A}+\gamma' \mathbf{I}]^{-1}\mathbf{B}\big\|_1 \\
      &\leq \big\|[\mathbf{A}+\gamma' \mathbf{I}]^{-1}\big\|_1 \big\|\mathbf{B}\big\|_1,
   \end{aligned}
   \end{equation}
   where $\big\|\cdot\big\|_1$ is the 1-norm (also referred to as the column-sum norm), which is simply the maximum absolute column sum of the matrix. According to the matrix theory, the following inequality holds:
   \begin{equation}\label{proof1.2}
   \begin{aligned}
     \big\|[\mathbf{A}+\gamma' \mathbf{I}]^{-1}\big\|_1 &\leq \sqrt{m} \big\|[\mathbf{A}+\gamma' \mathbf{I}]^{-1}\big\|_2 \\
     &= \sqrt{m} \lambda_{\max} \big[[\mathbf{A}+\gamma' \mathbf{I}]^{-1}\big],
   \end{aligned}
   \end{equation}
      where $\big\|\cdot\big\|_2$ is the 2-norm (also referred to as the spectral norm), which equals the maximum eigenvalue of the matrix denoted by $\lambda_{\max}[\cdot]$. Further, we have
   \begin{equation}\label{proof1.3}\small
   \begin{aligned}
       &\lambda_{\max}\big[[\mathbf{A}+\gamma' \mathbf{I}]^{-1}\big]\\
      =&\dfrac{1}{\lambda_{\min}\big[\mathbf{A}+\gamma' \mathbf{I}\big]}\\
      =&\dfrac{1}{\lambda_{\min}\Big[\sum\limits_{j = 1}^N {\sum\limits_{i = 1}^m {\dfrac{\lambda_i}{\sigma _i^2}{\kappa _{{\sigma _i}}}({e_j} - {c_i})}\bm{h}_j^T{\bm{h}_j}}\Big]+ \gamma'}\\
      \overset{(a)}{\leq}& \dfrac{1}{\dfrac{1}{\sigma^2}\lambda_{\min}\Big[\dfrac{\lambda_i}{\mu_i^2}{\kern-3pt}\sum\limits_{j=1}^{N}{\kern-3pt}\sum\limits_{i=1}^{m} \kappa_{\mu_i\sigma}\big(\delta\|\bm{h}_j\|_1+|t_j\!-\!c_i|\big)\bm{h}_j^T\bm{h}_j\Big] + \gamma'}\\
      =&\dfrac{\sigma^3\sqrt{2\pi}}{\lambda_{\min}(\theta)+\gamma''},
   \end{aligned}
   \end{equation}
   where $\gamma''=\sigma^3\sqrt{2\pi}\gamma'$, and (a) comes from
   \begin{equation}
   \begin{aligned}
      \big|e_j-c_i\big|&=\big|t_j-\bm{\beta}^T\bm{h}_j-c_i\big|\\
                   &\leq\big\|\bm{\beta}\big\|_1\big\|\bm{h}_j\big\|_1+\big|t_j-c_i\big|\\
                   &\leq\delta\big\|\bm{h}_j\big\|_1+\big|t_j-c_i\big|.
   \end{aligned}
   \end{equation}
   Likewise, we have
   \begin{equation}\label{proof1.4}
   \begin{aligned}
     \big\|\mathbf{B}\big\|_1&=\bigg\|\sum\limits_{j = 1}^{N} \sum\limits_{i = 1}^{m} \dfrac{\lambda_i}{\mu_i^2\sigma^2}{\kappa_{\mu_i\sigma}}\big(e_j \!-\! c_i\big)[t_j \!-\! c_i]\bm{h}_j^T\bigg\|_1\\
     &\overset{(b)}{\leq} \dfrac{1}{\sigma^3\sqrt{2\pi}}\sum\limits_{j = 1}^{N} \sum\limits_{i = 1}^{m} \dfrac{\lambda_i}{\mu_i^3}\big|t_j \!-\! c_i\big|\big\|\bm{h}_j^T\big\|_1,
   \end{aligned}
   \end{equation}
   where (b) is because $\kappa_{\sigma}(x) \leq \dfrac{1}{\sigma\sqrt{2\pi}}$ for any $x$.

   Combining (\ref{proof1.1})-(\ref{proof1.3}) and (\ref{proof1.4}), we have
   \begin{equation}
     \Big\|\bm{f}(\bm{\beta})\Big\|_1 \leq \dfrac{\vartheta\sqrt{m}}{\lambda_{\min}(\theta)+\gamma''}=\varphi(\sigma).
   \end{equation}

   Clearly, the function $\varphi(\sigma)$ is a continuous and monotonically decreasing function of $\sigma$ over $(0,\infty)$, satisfying $\lim\limits_{\sigma\to0}\varphi(\sigma)=\infty$, and $\lim\limits_{\sigma\to\infty}\varphi(\sigma)=\xi$. Therefore, if $\delta>\xi$, the equation $\varphi(\sigma)=\delta$ will have a unique solution $\sigma^*$ over $(0,\infty)$, and if $\sigma > \sigma^*$, we have $\varphi(\sigma) \leq \delta$, which completes the proof.
\end{IEEEproof}\vspace{-2ex}

\begin{theorem}
    If $\delta>\xi$ and $\sigma\geq max\{\sigma^*,\sigma^\dagger\}$, where $\sigma^*$ is the solution of equation $\varphi(\sigma)=\delta$, and $\sigma^\dagger$ is the solution of equation $\psi(\sigma)=\alpha (0<\alpha<1)$, where $\psi(\sigma)=\dfrac{\sqrt{m}(\Theta)}{(\lambda_{\min}(\theta)+\gamma'')\sigma^2}, \sigma\in(0,\infty)$ with $\Theta = \sum\limits_{j = 1}^{N} \sum\limits_{i = 1}^{m} \dfrac{\lambda_i}{\mu_i^5} \Big(\delta\big\|\bm{h}_j\big\|_1 \!+\! \big|t_j \!-\! c_i\big|\Big) \big\|\bm{h}_j\big\|_1 \Big(\delta\|\bm{h}_j^T\bm{h}_j^T\big\|_1 \!+\! \big|t_j\big|\big\|\bm{h}_j\big\|_1 \Big)$. Then it holds that $\big\|\bm{f}(\bm{\beta})\big\|_1 \leq \delta$ and $\big\|\nabla_{\bm{\beta}}\bm{f}(\bm{\beta})\big\|_1 \leq \alpha$ for all $ \bm{\beta} \in \{\bm{\beta} \in \mathbb{R}^L : \big\|\bm{\beta}\big\|_1 \leq \delta\}$.
\end{theorem}

\begin{IEEEproof}
   By Theorem 1, we have $\big\|\bm{f}(\bm{\beta})\big\|_1 \leq \delta$. To prove $\big\|\nabla_{\bm{\beta}}\bm{f}(\bm{\beta})\big\|_1 \leq \alpha$, it suffices to prove
   \begin{equation}
      \forall s, \big\|\dfrac{\partial}{\partial \beta_s}\bm{f}(\bm{\beta})\big\|_1 \leq \alpha,
   \end{equation}
   where
   \begin{small}
   \begin{equation}\label{proof2.1}
   \begin{aligned}
      &\Big\|\dfrac{\partial}{\partial \beta_s}\bm{f}(\bm{\beta})\Big\|_1\\
     =&\Big\|\dfrac{\partial}{\partial \beta_s}\big([\mathbf{A}+\lambda'I]^{-1}\mathbf{B}\big)\Big\|_1\\
     =&\bigg\|\!-\![\mathbf{A}+\lambda'I]^{-1}\Big(\dfrac{\partial}{\partial \beta_s}[\mathbf{A}+\lambda'I]\Big)[\mathbf{A}+\lambda'I]^{-1}\mathbf{B}\\
      &{\kern 1ex} + [\mathbf{A}+\lambda'I]^{-1}\Big(\dfrac{\partial}{\partial \beta_s}\mathbf{B}\Big)\bigg\|_1\\
         \leq&\Big\|[\mathbf{A}+\lambda'I]^{-1}\Big\|_1 \bigg\|\dfrac{\partial}{\partial \beta_s}[\mathbf{A}+\lambda'I]\bigg\|_1
              \Big\|\bm{f}(\bm{\beta})\Big\|_1\\
      &+ \Big\|[\mathbf{A}+\lambda'I]^{-1}\Big\|_1 \bigg\|\dfrac{\partial}{\partial \beta_s}\mathbf{B}\bigg\|_1.
   \end{aligned}
   \end{equation}
   \end{small}
   It is easy to derive
   \begin{equation}\label{proof2.2}
   \begin{aligned}
       &\bigg\|\dfrac{\partial}{\partial \beta_s}(\mathbf{A}+\lambda'I)\bigg\|_1\\
      =&\bigg\| \sum\limits_{j=1}^{N} \sum\limits_{i=1}^{m} \dfrac{\lambda_i}{\mu_i^4\sigma^4} (e_j-c_i) h_{js} {\kappa_{\sigma}\big(e_j-c_i\big)}\bm{h}_j{\bm{h}_j}^T\bigg\|_1\\
      \overset{(c)}{\leq}& \dfrac{1}{\sigma^5\sqrt{2\pi}} \sum\limits_{j=1}^{N} \sum\limits_{i=1}^{m} \dfrac{\lambda_i}{\mu_i^5} \Big(\delta\big\|\bm{h}_j\big\|_1+\big|t_j-c_i\big|\Big) \big\|\bm{h}_j\big\|_1 \big\|\bm{h}_j^T\bm{h}_j\big\|_1,
   \end{aligned}
   \end{equation}
   where (c) is due to the fact that $\big|(e_j-c_i)h_{js}\big| \leq \Big(\delta\big\|\bm{h}_j\big\|_1 + \big|t_j-c_i\big| \Big) \big\|\bm{h}_j\big\|_1$ and $\kappa_{\sigma}(x) \leq \dfrac{1}{\sigma\sqrt{2\pi}}$ for any $x$, in which $h_{is}$ is the $s$-th variable of $\bm{h}_i$. Similarly, one can derive
   \begin{small}
   \begin{equation}\label{proof2.3}
   \begin{aligned}
      \bigg\|\dfrac{\partial}{\partial \beta_s}\mathbf{B}\bigg\|_1
       {\kern -3pt} \leq {\kern -3pt} \dfrac{1}{\sqrt{2\pi}\sigma^5}{\kern -2pt}\sum\limits_{j = 1}^{N} \sum\limits_{i = 1}^{m} \dfrac{\lambda_i}{\mu_i^5} \Big(\delta\big\|\bm{h}_j\big\|_1 \!+\! \big|t_j \!-\! c_i\big|\Big) \big\|\bm{h}_j\big\|_1 \big\|t_j\bm{h}_j^T\big\|_1
   \end{aligned}
   \end{equation}
   \end{small}

   Then, combining (\ref{proof1.2}), (\ref{proof1.3}), (\ref{proof2.1})-(\ref{proof2.3})  and $\big\|\bm{f}(\bm{\beta})\big\|_1 \leq \delta$, we have
   \begin{equation}
        \Big\|\dfrac{\partial}{\partial \beta_s}\bm{f}(\bm{\beta})\Big\|_1 \leq \psi(\sigma).
   \end{equation}

   Obviously, $\psi(\sigma)$ is also a continuous and monotonically decreasing function of $\sigma$ over $(0,\infty)$, and satisfies $\lim\limits_{\sigma\to0}\psi(\sigma)=\infty$, and $\lim\limits_{\sigma\to\infty}\psi(\sigma)=0$. Therefore, given $0<\alpha<1$, the equation $\psi(\sigma)=\alpha$ has a unique solution $\sigma^\dagger$ over $(0,\infty)$, and if $\sigma > \sigma^\dagger$, we have $\psi(\sigma) \leq \delta$, which completes the proof.
\end{IEEEproof}

  According to Theorem 2 and contraction mapping theorem, given a certain parameter $\mu_i$ and an initial weight vector satisfying $\big\|\bm{\beta}_0\big\|_1 \leq \delta$, the FP-MMKCC algorithm will surely converge to a unique fixed point in the range $\bm{\beta} \in \{\bm{\beta} \in \mathbb{R}^L : \big\|\bm{\beta}\big\|_1 \leq \delta\}$ provided that the kernel bandwidth $\sigma$ is larger than a certain value. Moreover, since the $\alpha (0<\alpha<1)$ is the Lipschitz constant in the contraction mapping theorem, its value guarantees the convergence speed.
\end{appendices}

\ifCLASSOPTIONcaptionsoff
  \newpage
\fi

\bibliographystyle{IEEEtran}
\balance
\bibliography{MMKCC}

\begin{thebibliography}{10}
\providecommand{\url}[1]{#1}
\csname url@samestyle\endcsname
\providecommand{\newblock}{\relax}
\providecommand{\bibinfo}[2]{#2}
\providecommand{\BIBentrySTDinterwordspacing}{\spaceskip=0pt\relax}
\providecommand{\BIBentryALTinterwordstretchfactor}{4}
\providecommand{\BIBentryALTinterwordspacing}{\spaceskip=\fontdimen2\font plus
\BIBentryALTinterwordstretchfactor\fontdimen3\font minus
  \fontdimen4\font\relax}
\providecommand{\BIBforeignlanguage}[2]{{%
\expandafter\ifx\csname l@#1\endcsname\relax
\typeout{** WARNING: IEEEtran.bst: No hyphenation pattern has been}%
\typeout{** loaded for the language `#1'. Using the pattern for}%
\typeout{** the default language instead.}%
\else
\language=\csname l@#1\endcsname
\fi
#2}}
\providecommand{\BIBdecl}{\relax}
\BIBdecl

\bibitem{lin1990adaptive}
J.-H. Lin, T.~M. Sellke, and E.~J. Coyle, ``Adaptive stack filtering under the
  mean absolute error criterion,'' \emph{IEEE transactions on acoustics,
  speech, and signal processing}, vol.~38, no.~6, pp. 938--954, 1990.

\bibitem{coyle1988stack}
E.~J. Coyle and J.-H. Lin, ``Stack filters and the mean absolute error
  criterion,'' \emph{IEEE Transactions on Acoustics, Speech, and Signal
  Processing}, vol.~36, no.~8, pp. 1244--1254, 1988.

\bibitem{2pei1994least}
S.-C. Pei and C.-C. Tseng, ``Least mean p-power error criterion for adaptive
  fir filter,'' \emph{IEEE Journal on Selected Areas in Communications},
  vol.~12, no.~9, pp. 1540--1547, 1994.

\bibitem{liu2006error}
W.~Liu, P.~Pokharel, and J.~Principe, ``Error entropy, correntropy and
  m-estimation,'' in \emph{2006 16th IEEE Signal Processing Society Workshop on
  Machine Learning for Signal Processing}.\hskip 1em plus 0.5em minus
  0.4em\relax IEEE, 2006, pp. 179--184.

\bibitem{sayin2014novel}
M.~O. Sayin, N.~D. Vanli, and S.~S. Kozat, ``A novel family of adaptive
  filtering algorithms based on the logarithmic cost,'' \emph{IEEE Transactions
  on signal processing}, vol.~62, no.~17, pp. 4411--4424, 2014.

\bibitem{chen2012recursive}
X.~Chen, J.~Yang, J.~Liang, and Q.~Ye, ``Recursive robust least squares support
  vector regression based on maximum correntropy criterion,''
  \emph{Neurocomputing}, vol.~97, pp. 63--73, 2012.

\bibitem{13feng2015learning}
Y.~Feng, X.~Huang, L.~Shi, Y.~Yang, and J.~A. Suykens, ``Learning with the
  maximum correntropy criterion induced losses for regression.'' \emph{Journal
  of Machine Learning Research}, vol.~16, pp. 993--1034, 2015.

\bibitem{xu2018robust}
G.~Xu, B.-G. Hu, and J.~C. Principe, ``Robust c-loss kernel classifiers,''
  \emph{IEEE transactions on neural networks and learning systems}, vol.~29,
  no.~3, pp. 510--522, 2018.

\bibitem{syed2012correntropy}
M.~N. Syed, J.~C. Principe, and P.~M. Pardalos, ``Correntropy in data
  classification,'' in \emph{Dynamics of Information Systems: Mathematical
  Foundations}.\hskip 1em plus 0.5em minus 0.4em\relax Springer, 2012, pp.
  81--117.

\bibitem{singh2014c}
A.~Singh, R.~Pokharel, and J.~Principe, ``The c-loss function for pattern
  classification,'' \emph{Pattern Recognition}, vol.~47, no.~1, pp. 441--453,
  2014.

\bibitem{shi2018training}
W.~Shi, Y.~Gong, X.~Tao, and N.~Zheng, ``Training dcnn by combining max-margin,
  max-correlation objectives, and correntropy loss for multilabel image
  classification,'' \emph{IEEE transactions on neural networks and learning
  systems}, vol.~29, no.~7, pp. 2896--2908, 2018.

\bibitem{ren2020correntropy}
L.-R. Ren, Y.-L. Gao, J.-X. Liu, J.~Shang, and C.-H. Zheng, ``Correntropy
  induced loss based sparse robust graph regularized extreme learning machine
  for cancer classification,'' \emph{BMC bioinformatics}, vol.~21, no.~1, pp.
  1--22, 2020.

\bibitem{he2011robust}
R.~He, B.-G. Hu, W.-S. Zheng, and X.-W. Kong, ``Robust principal component
  analysis based on maximum correntropy criterion,'' \emph{IEEE Transactions on
  Image Processing}, vol.~20, no.~6, pp. 1485--1494, 2011.

\bibitem{zhou2017maximum}
N.~Zhou, Y.~Xu, H.~Cheng, Z.~Yuan, and B.~Chen, ``Maximum correntropy criterion
  based sparse subspace learning for unsupervised feature selection,''
  \emph{IEEE Transactions on Circuits and Systems for Video Technology}, 2017.

\bibitem{yu2020correntropy}
N.~Yu, M.-J. Wu, J.-X. Liu, C.-H. Zheng, and Y.~Xu, ``Correntropy-based
  hypergraph regularized nmf for clustering and feature selection on
  multi-cancer integrated data,'' \emph{IEEE Transactions on Cybernetics},
  2020.

\bibitem{zhao2011kernel}
S.~Zhao, B.~Chen, and J.~C. Principe, ``Kernel adaptive filtering with maximum
  correntropy criterion,'' in \emph{The 2011 International Joint Conference on
  Neural Networks}.\hskip 1em plus 0.5em minus 0.4em\relax IEEE, 2011, pp.
  2012--2017.

\bibitem{chen2016generalized}
B.~Chen, L.~Xing, H.~Zhao, N.~Zheng, and J.~C. Principe, ``Generalized
  correntropy adaptive filtering,'' \emph{IEEE Transactions on Signal
  Processing}, vol.~64, no.~13, pp. 3376--3387, 2016.

\bibitem{ma2015maximum}
W.~Ma, H.~Qu, G.~Gui, L.~Xu, J.~Zhao, and B.~Chen, ``Maximum correntropy
  criterion based sparse adaptive filtering algorithms for robust channel
  estimation under non-gaussian environments,'' \emph{Journal of the Franklin
  Institute}, vol. 352, no.~7, pp. 2708--2727, 2015.

\bibitem{wu2015robust}
Z.~Wu, S.~Peng, B.~Chen, and H.~Zhao, ``Robust hammerstein adaptive filtering
  under maximum correntropy criterion,'' \emph{Entropy}, vol.~17, no.~10, pp.
  7149--7166, 2015.

\bibitem{wu2015kernel}
Z.~Wu, J.~Shi, X.~Zhang, W.~Ma, and B.~Chen, ``Kernel recursive maximum
  correntropy,'' \emph{Signal Processing}, vol. 117, pp. 11--16, 2015.

\bibitem{cinar2012hidden}
G.~T. Cinar and J.~C. Pr{\'\i}ncipe, ``Hidden state estimation using the
  correntropy filter with fixed point update and adaptive kernel size,'' in
  \emph{Neural Networks (IJCNN), The 2012 International Joint Conference
  on}.\hskip 1em plus 0.5em minus 0.4em\relax IEEE, 2012, pp. 1--6.

\bibitem{liu2017maximum}
X.~Liu, B.~Chen, B.~Xu, Z.~Wu, and P.~Honeine, ``Maximum correntropy unscented
  filter,'' \emph{International Journal of Systems Science}, vol.~48, no.~8,
  pp. 1607--1615, 2017.

\bibitem{liu2007correntropy}
W.~Liu, P.~P. Pokharel, and J.~C. Pr{\'\i}ncipe, ``Correntropy: Properties and
  applications in non-gaussian signal processing,'' \emph{IEEE Transactions on
  Signal Processing}, vol.~55, no.~11, pp. 5286--5298, 2007.

\bibitem{chen2018mixture}
B.~Chen, X.~Wang, N.~Lu, S.~Wang, J.~Cao, and J.~Qin, ``Mixture correntropy for
  robust learning,'' \emph{Pattern Recognition}, vol.~79, pp. 318--327, 2018.

\bibitem{gonen2011multiple}
M.~G{\"o}nen and E.~Alpayd{\i}n, ``Multiple kernel learning algorithms,''
  \emph{Journal of machine learning research}, vol.~12, no. Jul, pp.
  2211--2268, 2011.

\bibitem{lanckriet2004learning}
G.~R. Lanckriet, N.~Cristianini, P.~Bartlett, L.~E. Ghaoui, and M.~I. Jordan,
  ``Learning the kernel matrix with semidefinite programming,'' \emph{Journal
  of Machine learning research}, vol.~5, no. Jan, pp. 27--72, 2004.

\bibitem{wang2007multik}
Z.~Wang, S.~Chen, and T.~Sun, ``Multik-mhks: a novel multiple kernel learning
  algorithm,'' \emph{IEEE Transactions on Pattern Analysis and Machine
  Intelligence}, vol.~30, no.~2, pp. 348--353, 2007.

\bibitem{yukawa2012multikernel}
M.~Yukawa, ``Multikernel adaptive filtering,'' \emph{IEEE Transactions on
  Signal Processing}, vol.~60, no.~9, pp. 4672--4682, 2012.

\bibitem{chen2017maximum}
B.~Chen, X.~Liu, H.~Zhao, and J.~C. Principe, ``Maximum correntropy kalman
  filter,'' \emph{Automatica}, vol.~76, pp. 70--77, 2017.

\bibitem{wang2017maximum}
F.~Wang, Y.~He, S.~Wang, and B.~Chen, ``Maximum total correntropy adaptive
  filtering against heavy-tailed noises,'' \emph{Signal Processing}, vol. 141,
  pp. 84--95, 2017.

\bibitem{principe2010information}
J.~C. Principe, \emph{Information theoretic learning: Renyi's entropy and
  kernel perspectives}.\hskip 1em plus 0.5em minus 0.4em\relax Springer Science
  \& Business Media, 2010.

\bibitem{santamaria2002adaptive}
I.~Santamar{\'\i}a, C.~Pantale{\'o}n, L.~Vielva, and J.~C. Principe, ``Adaptive
  blind equalization through quadratic pdf matching,'' in \emph{2002 11th
  European Signal Processing Conference}.\hskip 1em plus 0.5em minus
  0.4em\relax IEEE, 2002, pp. 1--4.

\bibitem{heravi2018new}
A.~R. Heravi and G.~A. Hodtani, ``A new information theoretic relation between
  minimum error entropy and maximum correntropy,'' \emph{IEEE Signal Processing
  Letters}, vol.~25, no.~7, pp. 921--925, 2018.

\bibitem{frank2010uci}
A.~Frank and A.~Asuncion, ``Uci machine learning repository [http://archive.
  ics. uci. edu/ml]. irvine, ca: University of california,'' \emph{School of
  information and computer science}.

\bibitem{wang2017stochastic}
D.~Wang and M.~Li, ``Stochastic configuration networks: Fundamentals and
  algorithms,'' \emph{IEEE transactions on cybernetics}, vol.~47, no.~10, pp.
  3466--3479, 2017.

\bibitem{wang2017robust}
------, ``Robust stochastic configuration networks with kernel density
  estimation for uncertain data regression,'' \emph{Information Sciences}, vol.
  412, pp. 210--222, 2017.

\bibitem{li2019robust}
M.~Li, C.~Huang, and D.~Wang, ``Robust stochastic configuration networks with
  maximum correntropy criterion for uncertain data regression,''
  \emph{Information Sciences}, vol. 473, pp. 73--86, 2019.

\bibitem{waibel1995phoneme}
A.~Waibel, T.~Hanazawa, G.~Hinton, K.~Shikano, and K.~J. Lang, ``Phoneme
  recognition using time-delay neural networks,'' \emph{Backpropagation:
  Theory, Architectures and Applications}, pp. 35--61, 1995.

\bibitem{kuo1993nonlinear}
J.-M. Kuo, ``Nonlinear dynamic modeling with artificial neural networks,''
  Ph.D. dissertation, Citeseer, 1993.

\bibitem{lotte2018review}
F.~Lotte, L.~Bougrain, A.~Cichocki, M.~Clerc, M.~Congedo, A.~Rakotomamonjy, and
  F.~Yger, ``A review of classification algorithms for eeg-based
  brain--computer interfaces: a 10 year update,'' \emph{Journal of neural
  engineering}, vol.~15, no.~3, p. 031005, 2018.

\bibitem{he2020brain}
B.~He, H.~Yuan, J.~Meng, and S.~Gao, ``Brain--computer interfaces,'' in
  \emph{Neural engineering}.\hskip 1em plus 0.5em minus 0.4em\relax Springer,
  2020, pp. 131--183.

\bibitem{islam2016methods}
M.~K. Islam, A.~Rastegarnia, and Z.~Yang, ``Methods for artifact detection and
  removal from scalp eeg: A review,'' \emph{Neurophysiologie Clinique/Clinical
  Neurophysiology}, vol.~46, no. 4-5, pp. 287--305, 2016.

\bibitem{tangermann2012review}
M.~Tangermann, K.-R. M{\"u}ller, A.~Aertsen, N.~Birbaumer, C.~Braun,
  C.~Brunner, R.~Leeb, C.~Mehring, K.~J. Miller, G.~Mueller-Putz \emph{et~al.},
  ``Review of the bci competition iv,'' \emph{Frontiers in neuroscience},
  vol.~6, p.~55, 2012.

\bibitem{lotte2010regularizing}
F.~Lotte and C.~Guan, ``Regularizing common spatial patterns to improve bci
  designs: unified theory and new algorithms,'' \emph{IEEE Transactions on
  biomedical Engineering}, vol.~58, no.~2, pp. 355--362, 2010.

\bibitem{rocha2013multiclass}
A.~Rocha and S.~K. Goldenstein, ``Multiclass from binary: Expanding
  one-versus-all, one-versus-one and ecoc-based approaches,'' \emph{IEEE
  Transactions on Neural Networks and Learning Systems}, vol.~25, no.~2, pp.
  289--302, 2013.

\bibitem{sun2010adaptive}
S.~Sun and R.~Huang, ``An adaptive k-nearest neighbor algorithm,'' in
  \emph{2010 seventh international conference on fuzzy systems and knowledge
  discovery}, vol.~1.\hskip 1em plus 0.5em minus 0.4em\relax IEEE, 2010, pp.
  91--94.

\bibitem{suykens1999least}
J.~A. Suykens and J.~Vandewalle, ``Least squares support vector machine
  classifiers,'' \emph{Neural processing letters}, vol.~9, no.~3, pp. 293--300,
  1999.

\bibitem{zhang2015convergence}
Y.~Zhang, B.~Chen, X.~Liu, Z.~Yuan, and J.~Principe, ``Convergence of a
  fixed-point minimum error entropy algorithm,'' \emph{Entropy}, vol.~17,
  no.~8, pp. 5549--5560, 2015.

\bibitem{chen2015convergence}
B.~Chen, J.~Wang, H.~Zhao, N.~Zheng, and J.~C. Principe, ``Convergence of a
  fixed-point algorithm under maximum correntropy criterion,'' \emph{IEEE
  Signal Processing Letters}, vol.~22, no.~10, pp. 1723--1727, 2015.

\end{thebibliography}

\end{document}